\documentclass[runningheads]{llncs}

\usepackage{eccv}

\usepackage{eccvabbrv}

\usepackage{graphicx}
\usepackage{booktabs}

\usepackage[accsupp]{axessibility}  

\usepackage[pagebackref,breaklinks,colorlinks,citecolor=eccvblue]{hyperref}

\usepackage{orcidlink}

\usepackage{amsmath}
\usepackage{amssymb}
\usepackage{siunitx}
\usepackage{multirow}
\usepackage{bm}

\begin{document}

\title{GHOST: \underline{G}rounded \underline{H}uman Motion Generation with \underline{O}pen Vocabulary \underline{S}cene-and-\underline{T}ext Contexts} 

\titlerunning{GHOST: \underline{G}rounded \underline{H}uman Motion Generation}

\author{Zolt\'{a}n~\'{A}.~Milacski\inst{1}\orcidlink{0000-0002-3135-2936} \and
Koichiro~Niinuma\inst{2}\orcidlink{0000-0001-8367-3988} \and
Ryosuke~Kawamura\inst{2}\orcidlink{0000-0001-5133-9838} \and
Fernando~de~la~Torre\inst{1}\orcidlink{0009-0007-3760-3007} \and
L\'{a}szl\'{o}~A.~Jeni\inst{1}\orcidlink{0000-0002-2830-700X}}

\authorrunning{Z.~\'{A}.~Milacski, K.~Niinuma et al.}

\institute{Robotics Institute, Carnegie Mellon University, Pittsburgh PA, USA 
\email{\{zmilacsk@andrew,ftorre@cs,laszlojeni@\}.cmu.edu}
\and
Fujitsu Research of America, Pittsburgh PA, USA
\email{\{kniinuma,k.ryosuke\}@fujitsu.com}}

\maketitle
\begin{figure}
\begin{center}
    \centering
    \captionsetup{type=figure}
  \centering
  \begin{subfigure}{0.35\linewidth}
    \includegraphics[width=0.95\linewidth]{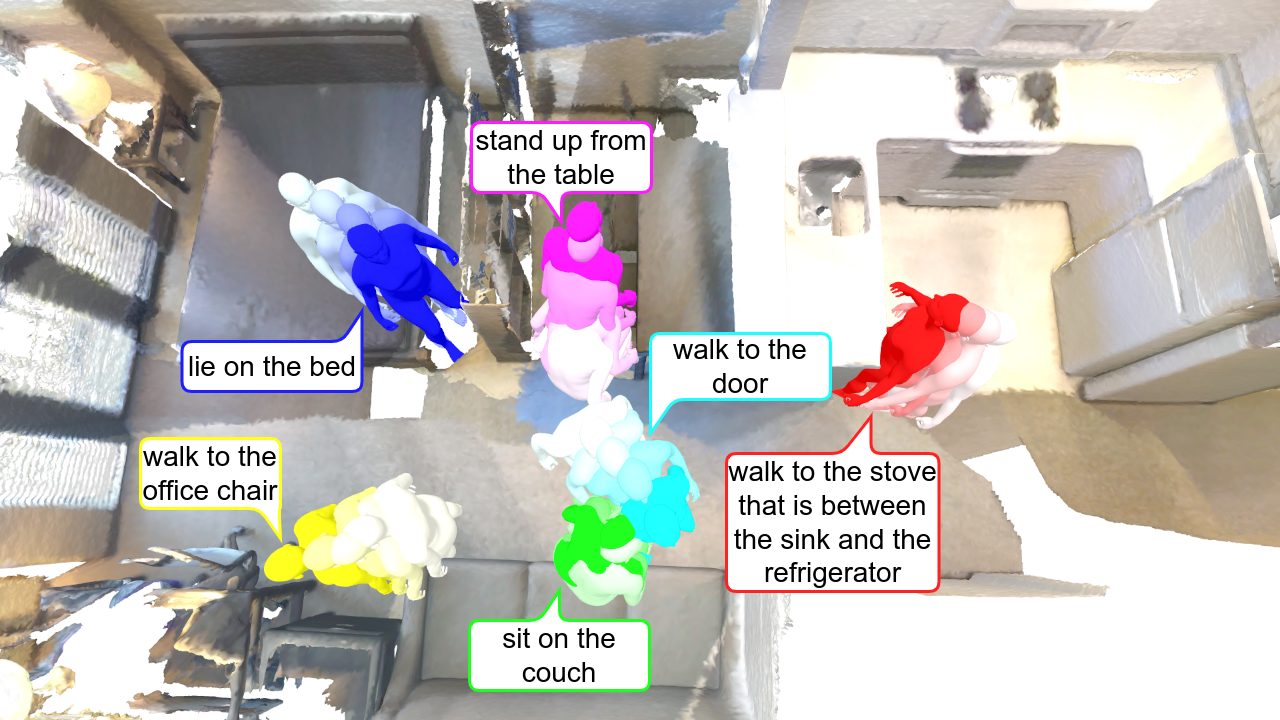}
    \caption{HUMANISE cVAE \cite{wang2022humanise}.}
    \label{fig:fancya}
  \end{subfigure}\hspace{-0.5em}%
  \begin{subfigure}{0.35\linewidth}
    \includegraphics[width=0.95\linewidth]{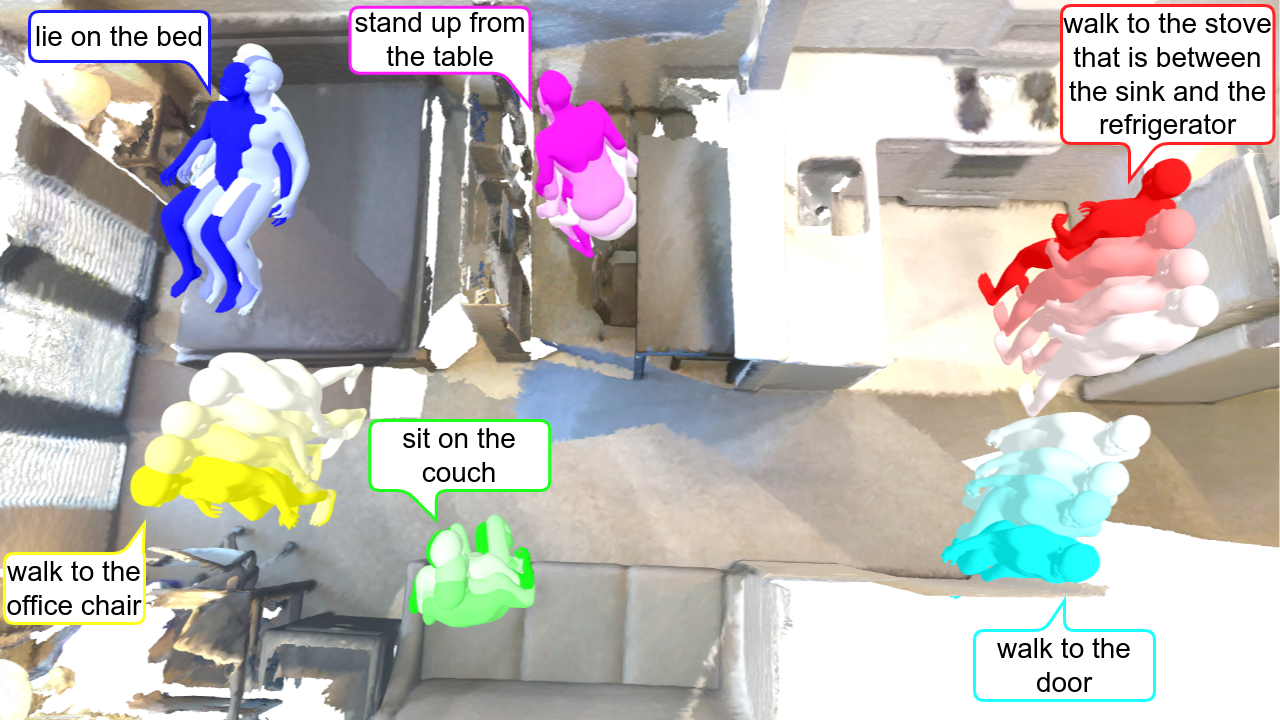}
    \caption{Our GHOST cVAE.}
    \label{fig:fancyb}
  \end{subfigure}\hspace{-0.5em}
  \begin{subfigure}{0.3\linewidth}
    \quad\quad\,\includegraphics[width=0.554\linewidth]{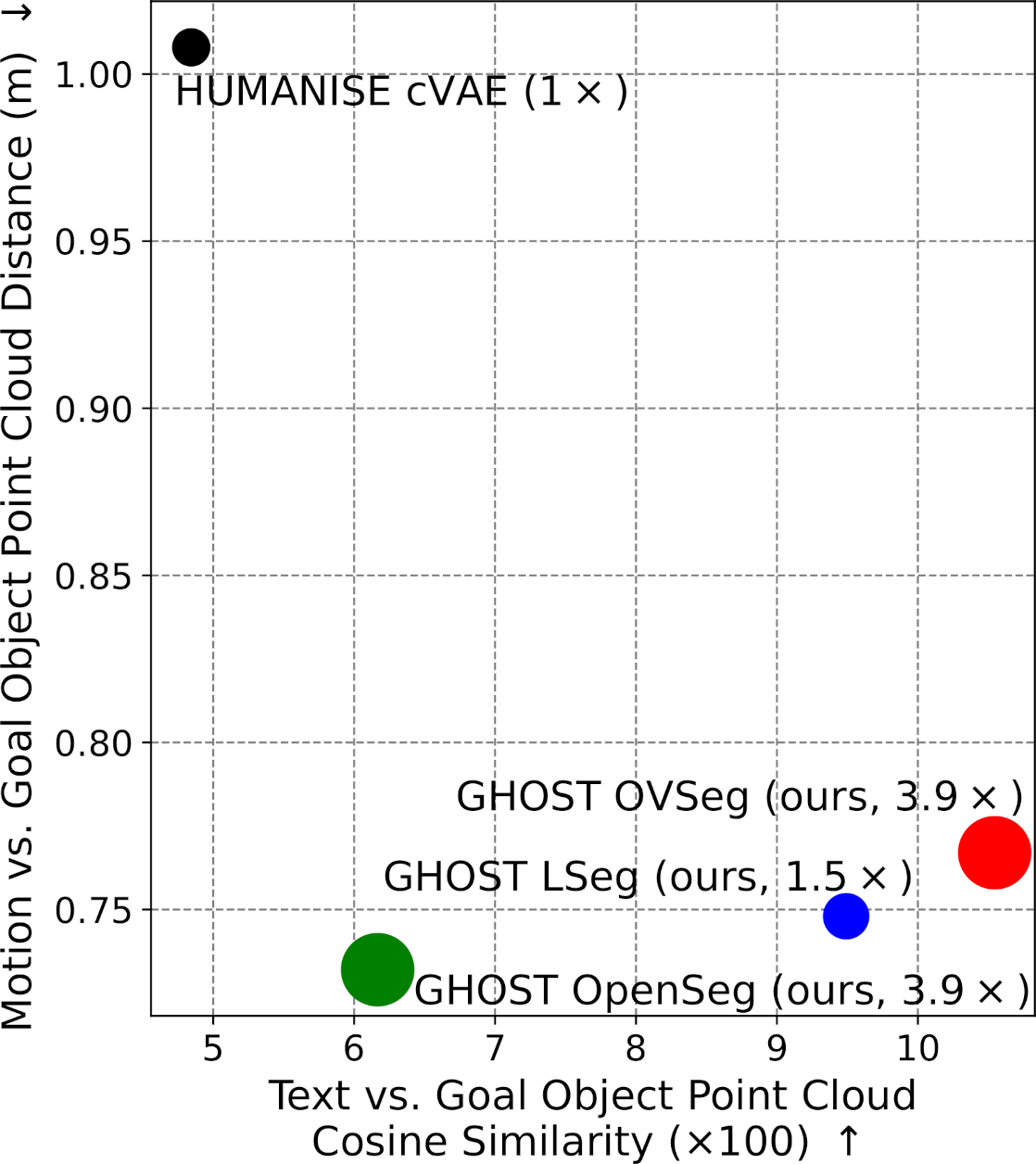}
    \caption{Comparison of models.}
    \label{fig:fancyc}
  \end{subfigure}
    \setcounter{figure}{0}
    \captionof{figure}{
    Comparison of our proposed GHOST cVAE method with the prior state-of-the-art HUMANISE cVAE \cite{wang2022humanise} in text-and-scene-conditional human motion generation.
    Best viewed in color.
    (a)~The HUMANISE cVAE exhibits a bias towards generating motions centered within the scene.
    (b)~In contrast, our GHOST cVAE demonstrates superior semantic understanding and achieves higher action performance.
    (c)~The three implementations of our GHOST framework exhibit approximately $1.5\times$ to $3.9\times$ larger parameter counts (indicated by dot radii) than the HUMANISE cVAE.
    All of our three variants outperform the baseline in two text-scene grounding metrics.
    }
    \label{fig:fancy}
\end{center}
\end{figure}

\begin{abstract}
  The connection between our 3D surroundings and the descriptive language that characterizes them would be well-suited for localizing and generating human motion in context but for one problem.
The complexity introduced by multiple modalities makes capturing this connection challenging with a fixed set of descriptors.
Specifically, closed vocabulary scene encoders, which require learning text-scene associations from scratch, have been favored in the literature, often resulting in inaccurate motion grounding.
In this paper, we propose a method that integrates an open vocabulary scene encoder into the architecture, establishing a robust connection between text and scene.
Our two-step approach starts with pretraining the scene encoder through knowledge distillation from an existing open vocabulary semantic image segmentation model, ensuring a shared text-scene feature space.
Subsequently, the scene encoder is fine-tuned for conditional motion generation, incorporating two novel regularization losses that regress the category and size of the goal object.
Our methodology achieves up to a $30\%$ reduction in the goal object distance metric compared to the prior state-of-the-art baseline model on the HUMANISE dataset.
This improvement is demonstrated through evaluations conducted using three implementations of our framework and a perceptual study.
Additionally, our method is designed to seamlessly accommodate future 2D segmentation methods that provide per-pixel text-aligned features for distillation.

\end{abstract}

\section{Introduction}
\label{sec:intro}

Human pose and motion generation in 3D scenes \cite{zhang2020generating, zhang2020generating2, starke2019neural, cao2020long, wang2021scene, hassan2021stochastic} plays a pivotal role in the realms of visual effects, video games, virtual and augmented reality, and robotics.
It empowers the creation of lifelike and expressive human animations within 3D environments, faithfully capturing spatial context and interactions.
By accounting for the geometry, lighting, and physical attributes of the 3D scene, human poses and motions can harmoniously meld with the environment, yielding immersive and visually cohesive animations.
Nevertheless, a significant limitation lies in the lack of precise control over the motion generation process, often relying on coarse assumptions regarding the location within the scene.
Simultaneously, recent strides in text-conditional generation have ushered in a revolution in synthetic data generation across a multitude of domains: images \cite{ramesh2021zero,frans2022clipdraw,ramesh2022hierarchical,saharia2022photorealistic,gal2022stylegan}, videos \cite{singer2022make, ho2022imagen}, 3D scenes \cite{poole2022dreamfusion}, 3D character shapes \cite{hong2022avatarclip,cao2023dreamavatar}, and human motion \cite{ahuja2019language2pose,ghosh2021synthesis,tevet2022motionclip, petrovich2022temos, TEACH:3DV:2022, kim2022flame, zhang2022motiondiffuse, tevet2022human}.
These advancements have paved the way for more intuitive and natural communication interfaces, enabling meticulous control over the generation process through the compositionality of language or even voice commands.
However, text-conditional motion generation methods often do not take into account any 3D scene context.
Bridging the gap between these modalities is essential to leverage both scene understanding and the precision of text-based control jointly, prompting the need for innovative motion synthesis approaches.

\begin{figure*}[t]
  \centering
  \begin{subfigure}{0.45\linewidth}
    \includegraphics[width=0.95\linewidth]{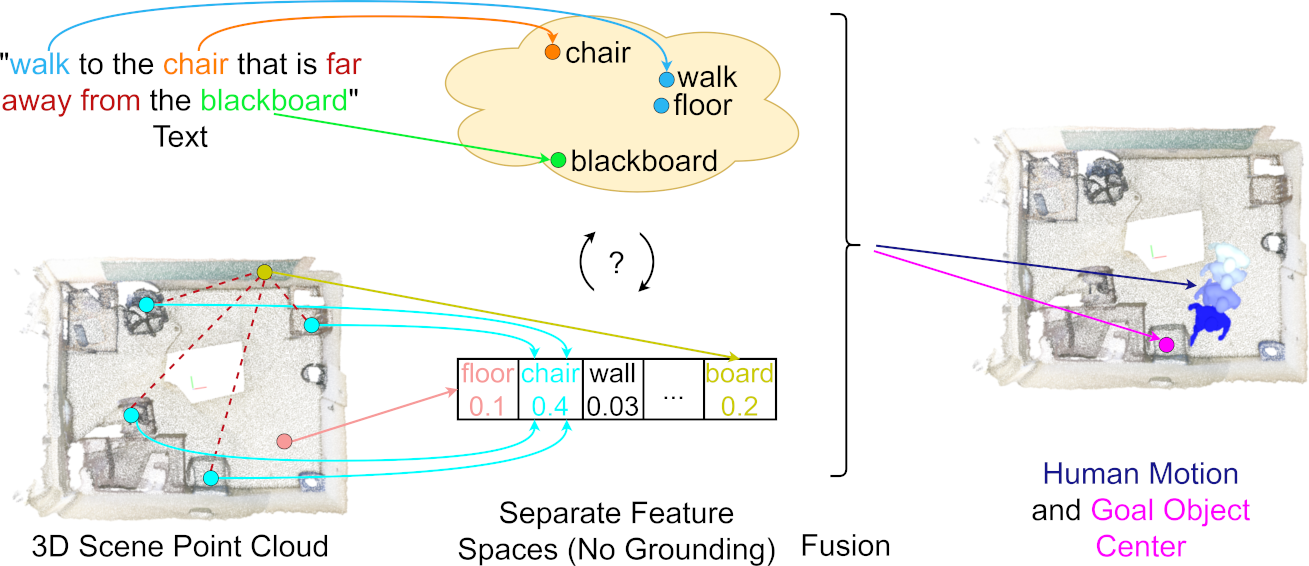}
    \caption{HUMANISE cVAE \cite{wang2022humanise}.}
    \label{fig:overview_idea_humanise}
  \end{subfigure}
  \quad
  \quad
  \begin{subfigure}{0.45\linewidth}
    \includegraphics[width=0.95\linewidth]{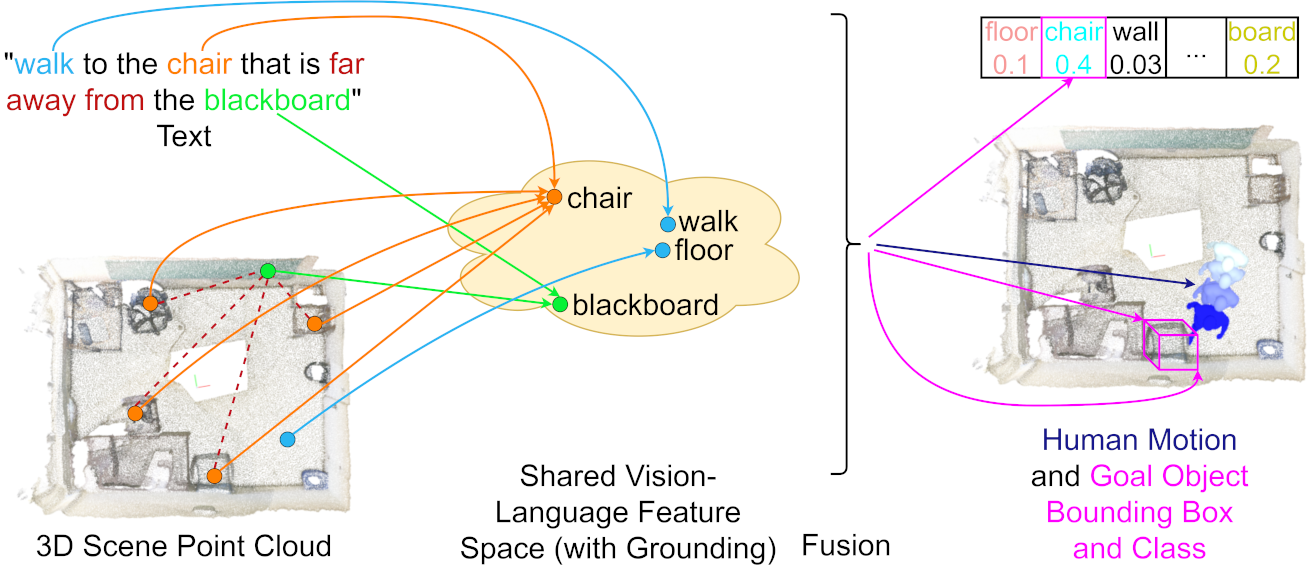}
    \caption{Our GHOST cVAE.}
    \label{fig:overview_idea_clipopulate}
  \end{subfigure}
  \caption{
  Overview of our idea.
  Best viewed in color.
  We compare our GHOST cVAE with the HUMANISE cVAE \cite{wang2022humanise} model. The major differences are in the text and 3D scene point cloud representations, grounding and regularization.
  (a)~The HUMANISE cVAE architecture utilizes a closed vocabulary scene encoder producing a finite set of labels, resulting in a misalignment with the open vocabulary text feature space.
  This requires the fusion module to learn grounding from scratch.
  Grounding is regularized by regressing the center point of the goal object.
  (b)~In contrast, our GHOST cVAE architecture employs a shared open vocabulary vision-language space for both modalities, establishing initial grounding between them.
  We regularize grounding by classifying and regressing the bounding box corners of the goal object, increasing awareness for category and size. 
  }
  \label{fig:overview_idea}
\end{figure*}

Recently, the HUMANISE \cite{wang2022humanise} dataset has been introduced for the task of text-and-scene-conditional human motion generation.
To the best of our knowledge, this is the only work towards this direction.
It comprises synthetic alignments of AMASS \cite{AMASS:ICCV:2019} motions with ScanNet \cite{dai2017scannet} scenes, as well as compositional template text annotations derived from BABEL \cite{BABEL:CVPR:2021} actions and Sr3D \cite{achlioptas2020referit3d} object referential utterances.
HUMANISE offers advantages over previous scene-conditional human motion datasets (PiGraphs \cite{savva2016pigraphs}, PROX-Qualitative \cite{PROX:2019}, GTA-IM \cite{cao2020long}), including larger size, greater scene diversity, consistent motion quality and semantic annotations.
Accompanying the HUMANISE dataset is a proposed Conditional Variational Autoencoder (cVAE) \cite{kingma2013auto,rezende2014stochastic,sohn2015learning} architecture that models the conditional probability of parameter sequences (global translation, global orientation, and body pose) for the SMPL-X \cite{SMPL-X:2019} human body model.
The condition module of the cVAE processes inputs from both text and scene point cloud modalities using their respective encoders and subsequent joint layers.
Interestingly, the authors in \cite{wang2022humanise} report an average distance of approximately $\SI{1}{\metre}$ from the goal object during motion sampling, while they also present qualitative failure cases where the character is positioned far away from the goal object, biased towards the center of the scene, as shown in \cref{fig:fancya}.
We argue that this limitation stems from employing a scene encoder that has been pretrained for closed vocabulary semantic segmentation, \ie, predicting a fixed set of categorical labels for each point.
This leads to a mismatch between the output spaces of the closed vocabulary scene encoder and the open vocabulary text encoder (depicted in \cref{fig:overview_idea_humanise}), where the latter is capable of embedding a broader and more diverse range of scenes and objects via natural language descriptions.
This compels the scene encoder to learn text-scene grounding from scratch on the dataset during fine-tuning for conditional motion generation.
Despite its recognition as the largest and most diverse available, the dataset falls short in meeting the demands of this task, resulting in improper grounding.

In this paper, we introduce \emph{GHOST}, an open vocabulary grounding framework designed to enhance text-and-scene-conditional human motion generation.
Our approach offers a two-step solution to circumvent the need for learning text-scene grounding from scratch, building upon recent advancements in open vocabulary scene segmentation methods \cite{peng2023openscene}.
Firstly, we establish a text-scene relationship before motion generation by leveraging the extensive grounding knowledge acquired by the Contrastive Language-Image Pretraining (CLIP) model \cite{radford2021learning} during its internet-scale vision-language pretraining.
This is achieved through pretraining a scene point cloud encoder, distilling knowledge from an Open Vocabulary Semantic Image Segmentation model on the ScanNet \cite{dai2017scannet} dataset.
Specifically, we create a correspondence between 3D scene points and text-aligned 2D viewpoint pixels in CLIP space, aligning our scene encoder's representations with those of the CLIP text encoder.
Secondly, similar to the original HUMANISE cVAE, we fuse the two modalities to train the conditional motion generator.
During this phase, we fine-tune the scene encoder with two novel auxiliary regularization losses strengthening the grounding of the goal object (\ie, regressing the bounding box coordinates and the ScanNet class).
The overview of our idea is presented in \cref{fig:overview_idea_clipopulate}.
We extensively evaluate the human motion grounding performance of three variants of our GHOST framework, each distilled from a distinct open vocabulary teacher model (LSeg \cite{li2022languagedriven}, OpenSeg \cite{ghiasi2022scaling}, and OVSeg \cite{liang2023open}), on the HUMANISE dataset through a comprehensive range of quantitative and qualitative experiments (see \cref{fig:fancyb}), including a perceptual user study.
Additionally, we conduct an ablation study to assess the individual effectiveness of each component of our model.\\
Our contributions can be summarized as follows:
\begin{itemize}
    \item We present \emph{GHOST}, a grounding framework for text-and-scene-conditional human motion generation.
    \item We establish a text-scene alignment in CLIP space, by replacing the closed vocabulary scene encoder pretraining with an open vocabulary knowledge distillation.
    \item We further refine the text-scene grounding by fine-tuning the scene encoder with two novel regularization losses that raise awareness to the category and the size of the goal object.
    \item We demonstrate substantially improved human motion placement performance during sampling on the HUMANISE dataset for all three tested teacher models.
\end{itemize}

\section{Related Work}
\subsection{Human Motion Generation}
Collecting large annotated datasets for human motion synthesis is challenging due to the need for motion capture and manual annotation.
As a result, supervised learning techniques are predominantly used for tasks like pose estimation from monocular images \cite{bogo2016keep,SMPL-X:2019}, videos \cite{kocabas2020vibe} or 2D poses \cite{chen20173d,martinez2017simple}, where massive amounts of paired input-motion data are available, necessitating modeling through deterministic (one-to-one) mappings.
In contrast, some approaches have explored unsupervised generative modeling on existing medium-sized motion datasets like AMASS \cite{AMASS:ICCV:2019}, focusing on capturing and sampling from the data distribution via stochastic architectures.
Unconditional motion generation aims to produce diverse and high-quality novel motion samples without specific constraints.
However, the lack of control over the generation restricts these methods to be used as motion priors \cite{SMPL-X:2019,kocabas2020vibe} or autoregressive motion prediction models \cite{barsoum2018hp,butepage2017deep,aliakbarian2020stochastic,yuan2020dlow}.

To address the need for both many-to-many mappings and increased control over sampling, conditional generation has garnered interest.
In the motion domain, condition-motion pairs are employed to train a stochastic model to generate diverse output motions for the same input condition or vice versa.
Various forms of motion conditioning have emerged, which can be categorized as follows.
\paragraph{Text-conditional generation.}
In this task, the condition is a text prompt, overcoming the constraints posed by the limited number of categories in class-conditional generation \cite{guo2020action2motion,cervantes2022implicit, petrovich2021action}, via leveraging the compositionality of natural language. Notable datasets for this problem include BABEL \cite{BABEL:CVPR:2021} and HumanML3D \cite{guo2022generating}. Various techniques have been proposed for this task, such as multimodal autoencoders \cite{ahuja2019language2pose,ghosh2021synthesis}, Conditional Variational Autoencoders (cVAEs) \cite{petrovich2022temos,TEACH:3DV:2022}, Conditional Generative Adversarial Networks (cGANs) \cite{Text2Action,lin2018human}, and Conditional Denoising Diffusion Probabilistic Models (cDDPMs) \cite{tevet2022human,zhang2022motiondiffuse,kim2022flame}.
\paragraph{Scene-conditional generation.}
These approaches, also known as Human-Scene Interaction (HSI) models, consider the environment when generating human motions. They account for scene layout, obstacles, spatial context, and object affordances.
Some methods focus on individual objects or actions, such as grasping \cite{chao2019visual,taheri2020grab,kim2014shape2pose}.
Others require additional input conditions, \eg, local motion \cite{hassan2021populating}, semantic segmentation \cite{zhang2020generating,zhang2020generating2}, or start and goal positions \cite{hassan2021stochastic,wang2021scene,wang2021synthesizing,starke2019neural}.
Moreover, certain approaches use test-time physical optimization \cite{yuan2023physdiff,huang2023diffusion} with scene constraints.
\paragraph{Text-and-scene-conditional generation.} This challenging problem involves leveraging both textual and scene conditions simultaneously, necessitating grounding between the two modalities.
The objective is to identify the \emph{goal object} among multiple instances of the same object class within complex 3D scenes, guided by textual descriptions of spatial relationships, and subsequently generate human motion to interact with the chosen object.
The only existing method in this category is a Conditional Variational Autoencoder (cVAE) architecture trained on the HUMANISE dataset \cite{wang2022humanise}.

In this paper, compared to HUMANISE, we demonstrate that substantial improvements can be achieved in grounding and human motion sampling performance by aligning the modalities in CLIP space via open vocabulary knowledge distillation and additional goal object regularization.

\subsection{Vision-Language Models and  Open Vocabulary Understanding}
\label{ss:vlmopen}
Vision-Language Models (VLMs) \cite{du2022survey,radford2021learning,jia2021scaling} have emerged as powerful grounding tools, bridging the gap between 2D visual and text modalities by mapping them to a shared feature space.
Open Vocabulary Understanding \cite{zareian2021open} denotes a model's capability to be queried with natural language prompts, facilitating the segmentation of 2D images and 3D scenes into their respective components.
This approach empowers the model to operate without being constrained by a predefined set of semantic categories or labels, fostering the recognition and understanding of a diverse array of objects and their associated properties.
\paragraph{Contrastive Language-Image Pretraining (CLIP).}
One notable VLM is CLIP \cite{radford2021learning}, which aligns the latent spaces of images and texts using contrastive learning.
Training on internet-scale paired data allows for a plethora of zero-shot text-controlled applications, \eg, image retrieval \cite{sain2023clip}, semantic image editing \cite{avrahami2022blended}, and 3D generation \cite{poole2022dreamfusion}.
CLIP is poor in encoding spatial relationships, as the texts primarily focus on identifying the foreground object.
\paragraph{Open Vocabulary Image Segmentation.}
Semantic image segmentation \cite{everingham2010pascal} algorithms often learn in supervised manner with closed set categories, and thus are unable to recognize more general concepts.
To address this limitation, LSeg \cite{li2022languagedriven} aligns dense pixel-level features with the CLIP text embedding of the associated pixel class name.
In a different approach, OpenSeg \cite{ghiasi2022scaling} aligns class-agnostic mask proposal features with individual words of a global image caption.
In contrast, OVSeg \cite{liang2023open} decouples mask proposal generation and open vocabulary classification into two distinct stages and fine-tunes CLIP for masked images.
\paragraph{Open Vocabulary 3D Scene Understanding.}
Traditional 3D scene understanding approaches employ task-specific supervised learning with ground truth 3D \cite{choy20194d,han2020occuseg,qi2017pointnet++} or 2D \cite{genova2021learning} labels, limiting scalability and generalization to diverse scenes.
To overcome these drawbacks, OpenScene~\cite{peng2023openscene} trains a 3D point cloud encoder by distilling 2D per-pixel LSeg or OpenSeg features through multi-view fusion, leading to zero-shot tasks like open vocabulary 3D segmentation, object affordance estimation, and 3D object search.

In this paper, we pretrain a point cloud encoder with the OpenScene loss function to achieve multimodal alignment with the CLIP text encoder.
Different from their approach, we fine-tune the scene encoder for text-and-scene-conditional human motion generation, with regularization to further refine grounding and spatial arrangement.

\section{Methods}
\subsection{Problem Definition and Notations}
Our goal is to populate 3D scenes with virtual 3D human motions via textual control.
Specificially, we aim to model the conditional probability $p\left(\bm{\Theta} \,|\, \bm{L}, \bm{S}\right)$, where $\bm{\Theta}=\{\bm{t},\bm{r},\bm{\theta}\}\in\mathbb{R}^{T \times (3+6+J\cdot 3)}$ denotes a sequence of human motion parameters (global translation $\bm{t}$, global orientation $\bm{r}$, body pose $\bm{\theta}$) of length $T$, $\bm{L}\in\mathbb{Z}^{W \times V}$ is a tokenized language description of length $W$ and vocabulary size $V$, and $\bm{S}\in\mathbb{R}^{N \times 6}$ is an RGB-colored scene point cloud.

We further use the differentiable SMPL-X \cite{SMPL-X:2019} body model to obtain human meshes for each motion frame, $\bm{\mathcal{M}}_t=\mathcal{M}(\bm{\Theta}_t,\bm{\beta})\in\mathbb{R}^{\SI{10475} \times 3}$, where $\mathcal{M}$ is linear blend skinning and $\bm{\beta}\in\mathbb{R}^{10}$ is the body shape.

\begin{figure*}[!t]
  \centering
  \begin{subfigure}{0.3443\linewidth}
\includegraphics[width=0.95\linewidth]{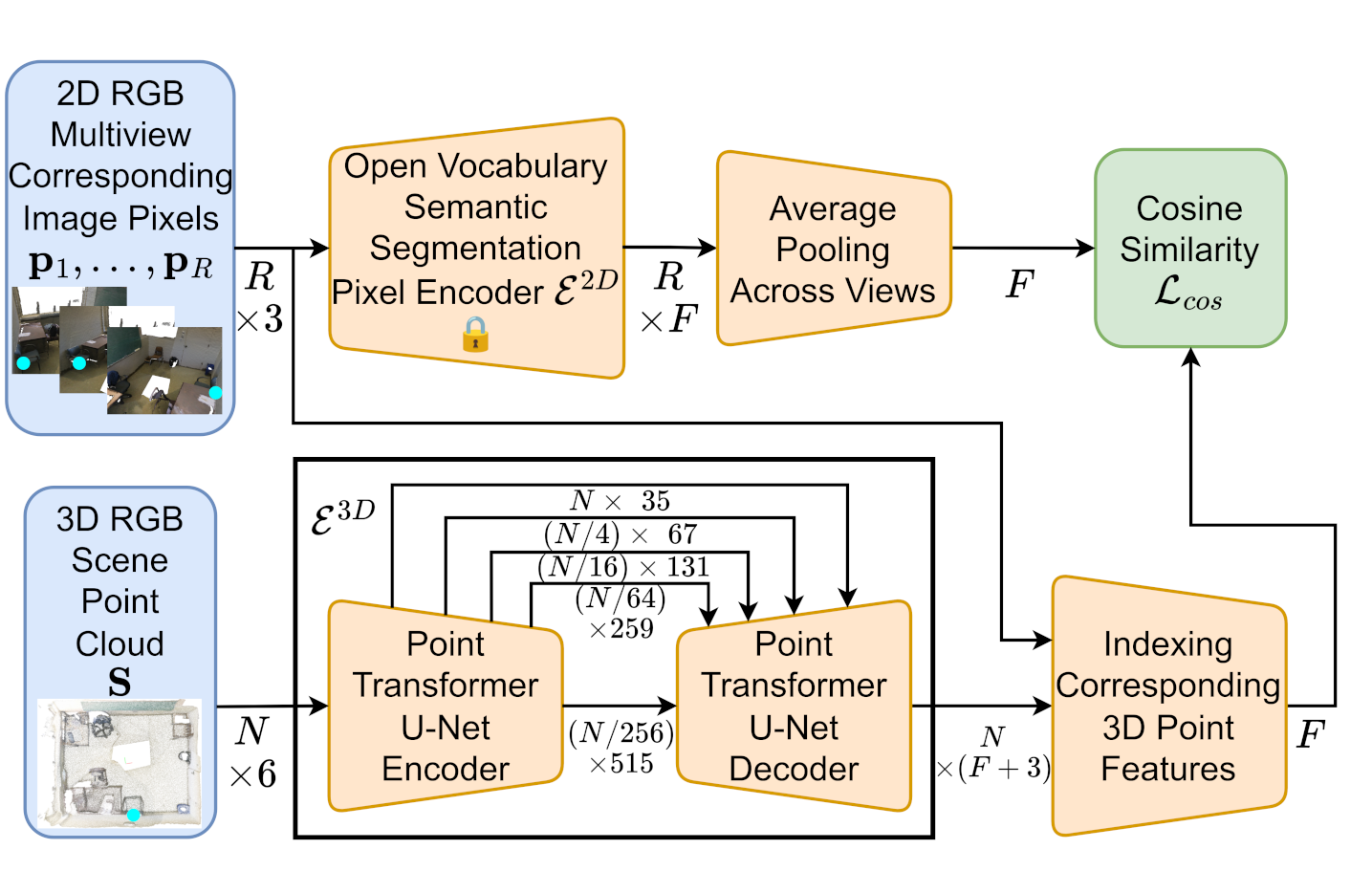}
    \caption{Pretraining.}
\label{fig:schem_diag_pretrain}
  \end{subfigure}
  \quad
  \quad
  \begin{subfigure}{0.5906\linewidth}
\includegraphics[width=0.95\linewidth]{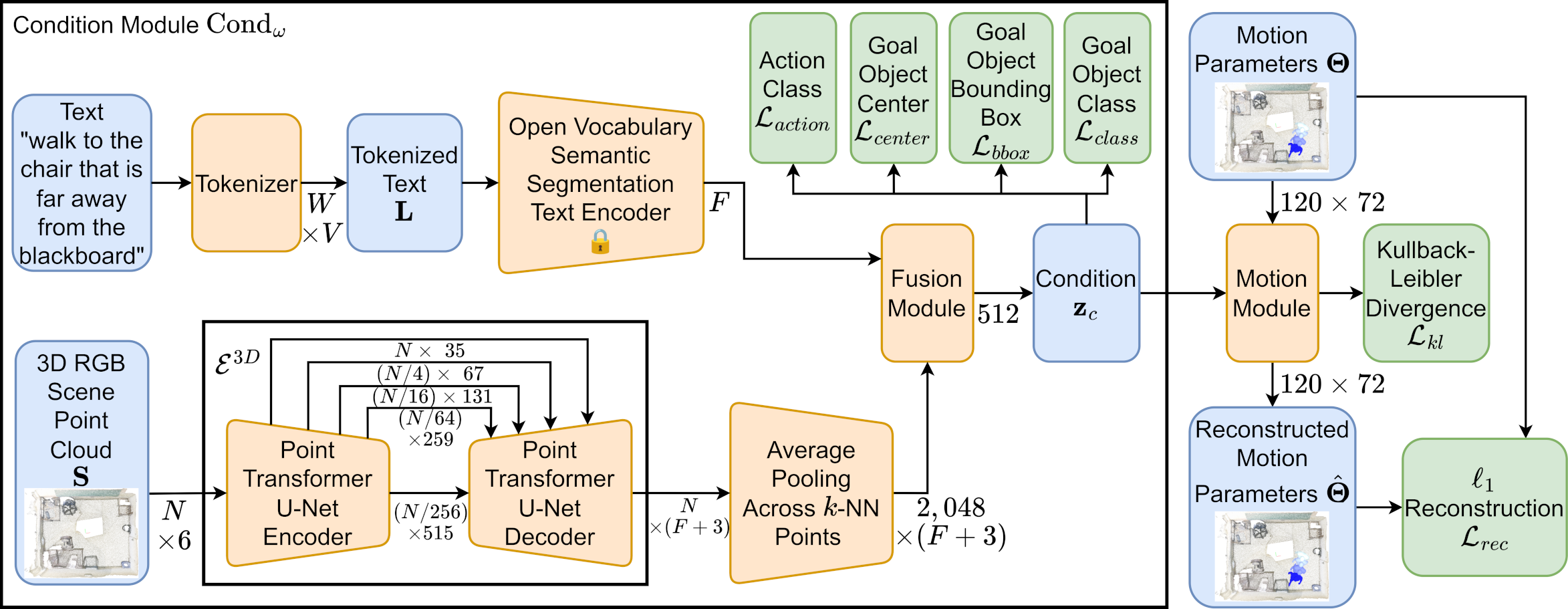}
    \caption{Training.}
\label{fig:schem_diag_train}
  \end{subfigure}
   \caption{
   Schematic diagram of the pretraining and training phases of our proposed GHOST framework for text-and-scene-conditional human motion generation.
   (a)~Pretraining involves maximizing the cosine similarity between our scene point cloud encoder and corresponding text-aligned 2D viewpoint pixel features, computed by an open vocabulary image segmentation teacher model.
   This ensures that our features align with text embeddings in a shared space.
   We use a Point Transformer U-Net scene encoder.
   (b)~Training employs a Conditional Variational Autoencoder (cVAE) architecture for motion generation, conditioned on both text and scene encoder outputs.
   The pretrained scene encoder weights are fine-tuned with two novel regularization losses (goal object bounding box regression and classification) to improve grounding.
   The rest of the components of the model remains consistent with the original HUMANISE cVAE \cite{wang2022humanise} model.
   }
   \label{fig:schem_diag}
\end{figure*}

\subsection{Proposed Solution}
To tackle the task, we introduce a cVAE generative model to capture the desired conditional probability, as shown in \cref{fig:schem_diag}.
While we largely adopt the motion module of the HUMANISE cVAE \cite{wang2022humanise}, our contribution lies in improved text-scene grounding through open vocabulary pretraining.
\paragraph{Motion Module.}
The motion module architecture is identical to its equivalent from the vanilla HUMANISE cVAE \cite{wang2022humanise}.
It takes the motion $\bm{\Theta}$ and the condition $\bm{z}_c\in\mathbb{R}^{C}$ as input, and outputs a reconstructed motion $\hat{\bm{\Theta}}$.

The motion encoder $\mathrm{Enc}_{\bm{\psi}}$ consists of a bidirectional GRU \cite{cho2014learning} layer, concatenation with $\bm{z}_c$, a residual block, and linear output layers for the mean and covariance parameters of the Gaussian distribution ($\bm{\mu}\in\mathbb{R}^{Z}$ and $\bm{\Sigma}\in\mathbb{R}^{Z \times Z}$).
The reparametrization trick \cite{kingma2013auto} is then employed to sample a latent variable $\bm{z}\in\mathbb{R}^{Z}$.
Finally, the motion decoder $\mathrm{Dec}_{\bm{\phi}}$ combines $\bm{z}$ and $\bm{z}_c$ using a linear layer, and utilizes a sinusoidal positional embedding, a transformer decoder \cite{vaswani2017attention}, and a linear output layer.
\paragraph{Condition Module.}
Here, we present our condition module $\mathrm{Cond}_{\bm{\omega}}$.
Compared to the HUMANISE cVAE \cite{wang2022humanise}, we apply different text and scene encoder architectures.
$\mathrm{Cond}_{\bm{\omega}}$ takes the tokenized text $\bm{L}$ and the scene point cloud $\bm{S}$ as input, and outputs the condition $\bm{z}_c=\mathrm{Cond}_{\bm{\omega}}(\bm{S},\bm{L})$.
The architecture is summarized in \cref{fig:schem_diag_train}.

We introduce separate text and scene encoders dedicated to processing $\bm{L}$ and $\bm{S}$, along with a fusion module for combining their codes.
Our text encoder is that of an open vocabulary image segmentation model (see \cref{ss:vlmopen,ss:models}).
For the scene encoder, we adopt a Point Transformer (PT) \cite{zhao2021point} to compute features for each point jointly,  and a downsampling module.
Different from HUMANISE cVAE, which employs solely an encoder, we utilize an entire U-Net \cite{ronneberger2015u} architecture $\mathcal{E}^{3D}$ (encoder and decoder with residual skip connections).
The downsampling module involves farthest point sampling \cite{qi2017pointnet++} and average pooling across $k$-nearest points.
Finally, to fuse the text and scene features, we concatenate them and apply a Self-Attention \cite{lin2017structured} layer.
The resulting point features and coordinates are passed through dense ReLU and linear layers to obtain the fused scene feature.
Finally, the fused scene and text features are concatenated with the SMPL-X shape $\bm{\beta}$ and transformed by a linear layer to get the conditional latent $\bm{z}_c$.
\paragraph{Pretraining.}
In contrast to HUMANISE cVAE, which utilizes a PT scene encoder pretrained for closed vocabulary semantic segmentation, we distill our U-Net student model $\mathcal{E}^{3D}$ using the open vocabulary OpenScene loss \cite{peng2023openscene}:
\begin{equation}
    \mathcal{L}_{cos}=1-\cos\left(\frac{1}{R}\sum_{j=1}^{R}{\left[\mathcal{E}^{2D}(\bm{I}_j)\right]_{(\bm{S}_{\cdot,:3}\bm{P}_{j}),\cdot}},\mathcal{E}^{3D}(\bm{S})\right),\\
\end{equation}
\ie, we maximize the cosine similarity between text-aligned 2D pixel features and our U-Net output.
Here, $\mathcal{E}^{2D}$ represents the per-pixel encoder of an open vocabulary image segmentation model with feature size $F$ (see \cref{ss:vlmopen} and \cite{peng2023openscene}) that we use as the teacher, $\bm{I}_j\in\mathbb{R}^{H \times W}$ is the $j$th of $R$ 2D viewpoint images, and $\bm{P}_j\in\mathbb{R}^{3 \times 2}$ is the $j$th view projection matrix.
As opposed to OpenScene \cite{peng2023openscene}, which trains a MinkowskiNet \cite{choy20194d} with this loss, we use a PT architecture.
Various open vocabulary segmentation models can be seamlessly integrated as the teacher, offering a plug-and-play framework.
To establish an alignment, we adopt and freeze the text encoder parameters of the teacher model.
\cref{fig:schem_diag_pretrain} provides an overview of this phase.
\paragraph{Training.}
Our loss function is similar to the one proposed for the vanilla HUMANISE cVAE \cite{wang2022humanise}, but we incorporate two novel terms.
Our overall objective is:
\begin{align}
\begin{split}
    \mathcal{L} &= \mathcal{L}_{rec}
    +\mathcal{L}_{reg}, \\
    \mathcal{L}_{rec} &= \mathcal{L}_{\bm{t}}
    + \lambda_{\bm{r}} \mathcal{L}_{\bm{r}}
    + \lambda_{\bm{\theta}} \mathcal{L}_{\bm{\theta}}
    + \lambda_{\bm{\mathcal{M}}} \mathcal{L}_{\bm{\mathcal{M}}}, \\
    \mathcal{L}_{reg} &= \lambda_{kl} \mathcal{L}_{kl}
    +\lambda_{action} \mathcal{L}_{action}
    +\lambda_{center} \mathcal{L}_{center} 
    +\lambda_{bbox} \mathcal{L}_{bbox}
    +\lambda_{class} \mathcal{L}_{class},
\end{split}
\end{align}
where $\mathcal{L}_{rec}$ is an $\ell_1$ reconstruction loss between true and predicted SMPL-X parameters ($\mathcal{L}_{\bm{t}}$ for global translation, $\mathcal{L}_{\bm{r}}$ for global orientation, $\mathcal{L}_{\bm{\theta}}$ for body pose) and canonical mesh vertices ($\mathcal{L}_{\bm{\mathcal{M}}}$).
$\mathcal{L}_{reg}$ is a regularization loss consisting of a Kullback\,--\,Leibler divergence term $\mathcal{L}_{kl}=D_{KL}\left[\mathcal{N}(\bm{\mu},\bm{\Sigma})\, \| \,\mathcal{N}(\bm{0},\bm{I})\right]$ promoting a standard Gaussian latent $\bm{z}$, along with four grounding loss terms.
These include auxiliary linear regressors on top of the condition $\bm{z}_c$, designed to improve awareness of the action and the goal object.
Similar to HUMANISE \cite{wang2022humanise}, we employ $\mathcal{L}_{action}$, a cross-entropy loss for action classification, and $\mathcal{L}_{center}$, the mean squared error for goal object center coordinates.
Additionally, we introduce two novel losses: $\mathcal{L}_{bbox}$, the mean squared error for goal object bounding box corner coordinates (axis aligned), and $\mathcal{L}_{class}$, a cross-entropy loss for goal object classification with $9$ ScanNet \cite{dai2017scannet} categories.
\cref{fig:schem_diag_train} illustrates this stage.

Different from OpenScene \cite{peng2023openscene}, which addresses zero-shot tasks with fixed scene and text encoders, we fine-tune the scene encoder for conditional motion generation while keeping the text encoder frozen.
This allows our model to capture 3D spatial relationships (``where''), addressing a shared weakness of CLIP, open vocabulary segmentation methods, and OpenScene that primarily focus on ``what''.

\section{Experimental Setup}
\subsection{Dataset}
To evaluate our hypotheses, we conducted experiments using the HUMANISE \cite{wang2022humanise} dataset, following their recommended settings to ensure a fair comparison.

The HUMANISE dataset comprises \SI{19648}{} AMASS \cite{AMASS:ICCV:2019} motion sequences that have been synthetically aligned with 643 ScanNet \cite{dai2017scannet} scenes, resulting in a comprehensive collection of 1.2 million motion frames.
The motion sequences correspond to four distinct BABEL \cite{BABEL:CVPR:2021} actions: walk (\SI{8264}{}), sit (\SI{5578}{}), stand up (\SI{3463}{}), and lie (\SI{2343}{}).
The motions were encoded using the parameter sequences of the gender neutral SMPL-X body model with $J=21$ joints, resulting in a total of $72$ parameters per frame.
For batch processing, we padded motion sequences to a fixed length of $T=120$.
For encoding the ScanNet scenes, we randomly sampled $N=\SI{32768}{}$ vertices from the scanned scene mesh.
The text annotations for the motions follow the template-based format of Sr3D \cite{achlioptas2020referit3d}, including an optional spatial relation with nearby ``anchor'' objects:
{
\small{``\texttt{\text{$\langle$action$\rangle$}}}
\small{\texttt{\text{$\langle$goal object class$\rangle$}}}
\small{\texttt{\text{$\bigl[\langle$spatial relation$\rangle$}}}
\small{\texttt{\text{$\langle$anchor object classes$\rangle \bigr]$}}''}.
}
We augmented the dataset by applying random rotations and translations to each scene-motion pair.
We utilized the official training-test set split for each action subset.

\subsection{Hyperparameter Settings}
Throughout our experiments, we adhere to specific hyperparameter settings for consistency and fair comparison, many of which are adopted from \cite{wang2022humanise}.
The following list provides an overview of these hyperparameters.

We train all models over 150 epochs using the Adam \cite{kingma2014adam} optimizer with a learning rate of $10^{-4}$ and a batch size of 24.
Additionally, we manually tuned the regularization parameters, setting $\lambda_{kl}=\lambda_{center}=\lambda_{bbox}=0.1$, $\lambda_{action}=\lambda_{class}=0.5$, $\lambda_{\bm{r}}=1.0$ and  $\lambda_{\bm{\theta}}=\lambda_{\bm{\mathcal{M}}}=10.0$.

The cVAE hyperparameters can be summarized as listed below.
The bidirectional GRU text encoder layer has 256 units, the size of the latent $\bm{z}$ is $Z=256$, and the transformer decoder layer has 512 units.
Next, we present the PT U-Net architecture $\mathcal{E}^{3D}$ of the scene encoder.
It comprises 5 encoder stages, each consisting of a transition down module and a varying number of PT Blocks (2, 3, 4, 6 and 3, respectively).
The decoder component contains 5 stages with a transition up module and 2 PT Blocks in each.
The output head includes a ReLU activation and a linear layer with $F$ units.
Each PT Block incorporates a Self-Attention layer, linear projections, and a residual skip connection.
During fine-tuning, we use a learning rate of $10^{-5}$.
Downsampling involves reducing the number of points from \SI{32768}{} to \SI{2048}{} and averaging features across $k=16$ nearest neighbors.
The dimensionality of the condition $\bm{z}_c$ is $C=512$.

We use an NVIDIA\textsuperscript\textregistered\,A100 \SI{80}{\giga\byte} GPU for training.

\subsection{Teacher Models, Baseline and Ablation}
\label{ss:models}
We tested three implementations of our framework, each distilled from a different open vocabulary image segmentation teacher model $\mathcal{E}^{2D}$ (LSeg \cite{li2022languagedriven}, OpenSeg \cite{ghiasi2022scaling}, and OVSeg \cite{liang2023open}, see \cref{ss:vlmopen}), complemented by the corresponding CLIP text encoder (\mbox{ViT-B/32}, \mbox{ViT-L/14@336px}, and \mbox{ViT-L/14} with $F\in\{512,768,$ $768\}$, respectively, with $W=77$, $V=\SI{49407}{}$).

To evaluate the effectiveness, we conducted a comparison against the unmodified HUMANISE cVAE \cite{wang2022humanise}.
To the best of our knowledge, this is the only existing competing method.
It can be regarded as a strong baseline, as the authors have thoroughly explored diverse design choices (scene encoders, regularizers, and multimodal fusion strategies) to optimize their architecture.

We also performed an ablation study to assess the impact of specific components in our framework.
We evaluated our approach against four simplified variants, where we replaced either the text or the scene encoder with its counterpart from the vanilla HUMANISE cVAE, or we set either $\lambda_{bbox}=0$ or $\lambda_{class}=0$ for our proposed loss terms.
We performed ablation on the walk action subset due to computational constraints.
This subset is big enough for statistical significance, yet computationally cheaper than the entire HUMANISE dataset.

\subsection{Evaluation Metrics}
We assess model performance using a set of quantitative evaluation metrics.
As our main objective is to enhance grounding, we focus primarily on the distance between the generated motion and the goal object, along with perceptual quality.
For additional metrics concerning motion reconstruction quality and diversity, we kindly refer the reader to the supplementary material.
It is important to note that motion reconstruction is an easier task as it has access to the ground truth location, thus, we place less emphasis on it.
\paragraph{Generation.}
Along with identifying the goal object from text, one should generate a human motion that is close enough to interact with it.
To quantify the effectiveness of this capability, we measure the mean distance between $K$ generated humans and the goal object, which is computed as follows:
\begin{equation}
d(\bm{L}, \bm{S})
=
\frac{1}{K}\sum_{j=1}^{K}{\mathrm{ReLU}{\left[\min{\left(\mathrm{SDF}_{\hat{\bm{\mathcal{M}}}_{t}^{(j)}}^{+}\left[\bm{S}_{goal,:3}\right]\right)}\right]}},
\label{eq:humangoaldist}
\end{equation}
where 
$\hat{\bm{\mathcal{M}}}_t^{(j)}=\mathcal{M}\left[\bm{\hat{\Theta}}_t^{(j)},\bm{\beta}\right]=\mathcal{M}\left[\mathrm{Dec}_{\bm{\phi}}\left(\bm{z}^{(j)},\bm{z}_c\right)_t,\bm{\beta}\right]$ 
is the SMPL-X human body mesh sampled from random standard Gaussian latent $\bm{z}^{(j)}\sim\mathcal{N}(\bm{0},\bm{I})$ and condition $\bm{z}_c$ at time step $t$, whereas $\mathrm{SDF}_{\hat{\bm{\mathcal{M}}}_t^{(j)}}^{+}[\cdot]$ is its positive Signed Distance Function (SDF).
We evaluate the SDF at $\bm{S}_{goal,:3}\in\mathbb{R}^{G \times 3}$, the subset of $\bm{S}$ corresponding to the goal object in text $\bm{L}$.
We identify the smallest distance, and if it is negative, we replace it with zero to disregard the penetration.
We use the last motion frame $t=T$ for walk, sit and lie; and the first frame $t=1$ for stand up; with $K=10$.
\paragraph{Perceptual Study.}
To gain insights into the quality and coherence of the generated motions from a perceptual standpoint, we conducted a Two-Alternative Forced Choice (2AFC) user study with 27 participants, each assessing 60 pairs of videos generated from 20 text-scene combinations.
For each pair, one video was generated by the HUMANISE cVAE baseline, and the other by our GHOST OpenSeg model, both trained on the entire HUMANISE dataset.
Pairs were shuffled randomly, and participants selected the video aligning better with the given textual description.
To aid the participants in identifying the ground truth goal object, we highlighted it in red color within the scene.

\section{Results}
\subsection{Quantitative Results}
\cref{tab:quant}, \cref{tab:quant_user} and \cref{tab:quant_ablation} present the quantitative generation results on the HUMANISE dataset, focusing on motion grounding quality, our perceptual study, and ablation analysis.

\begin{table}[!t]
\centering
\caption{
Quantitative results of generation experiments on the HUMANISE dataset.
The winning numbers are highlighted in bold for each action subset.}
\begin{tabular}{@{}lccccc@{}}
\toprule
& \multicolumn{5}{c}{Goal Object Distance ($\SI{}{\metre}$) $\downarrow$} \\
\cmidrule(lr){2-6} 
Method & walk & sit & stand up & lie & all \\
\midrule
HUMANISE cVAE \cite{wang2022humanise} & 1.370 & 0.903 & 0.802 & 0.196 & 1.008 \\
GHOST LSeg (ours) & 1.090 & 0.695 & 0.767 & \textbf{0.185} & 0.748\\
GHOST OpenSeg (ours) & \textbf{0.952} & \textbf{0.668} & \textbf{0.600} & 0.200 & \textbf{0.732} \\
GHOST OVSeg (ours) & 1.027 & 0.680 & 0.626 & 0.263 & 0.767\\
\bottomrule
\end{tabular}
\label{tab:quant}
\end{table}

\begin{table}[!t]
\centering
\caption{
Quantitative results of the perceptual study of agnostic all-actions models trained on the entire HUMANISE dataset.
The winning numbers are highlighted in bold.}
\begin{tabular}{@{}lcc@{}}
\toprule
& \# of Participants & Total Preference \\
Method &  Preferring $\uparrow$ & Percentage $\uparrow$ \\
\midrule
HUMANISE cVAE \cite{wang2022humanise} & \phantom{0}0 & 36.73\% \\
GHOST OpenSeg (ours) & \textbf{27} & \textbf{63.27}\textbf{\%} \\
\bottomrule
\end{tabular}
\label{tab:quant_user}
\end{table}

Regarding the goal distance $d(\bm{L},\bm{S})$ from \eqref{eq:humangoaldist} during sampling, our framework demonstrated significant improvements over the HUMANISE cVAE baseline across all three implementations, detailed in \cref{tab:quant}.
Our model with OpenSeg distillation outperformed others, achieving remarkable reductions of \SI{41.8}{\centi\metre} on the largest action-specific walk subset and \SI{27.6}{\centi\metre} on the entire dataset.
It only marginally trailed by \SI{0.4}{\centi\metre} in the smallest and easiest lie subset with larger goal objects, where our LSeg variant excelled the most.

The results of the perceptual study are shown in \cref{tab:quant_user}.
Our OpenSeg distilled method's samples were preferred over the baseline's samples 63.27\% of the time.
Furthermore, all 27 participants unanimously preferred the motions generated by our model.
These indicate that our approach achieved better alignment with the provided texts compared to the baseline method.

\begin{table}[!t]
\centering
\caption{
Quantitative results of ablation experiments on the walk action subset of the HUMANISE dataset.
The winning number is highlighted in bold.}
\begin{tabular}{@{}lc@{}}
\toprule
Method & Goal Obj. Dist. ($\SI{}{\metre}$) $\downarrow$ \\
\midrule
GHOST OpenSeg w. BERT \cite{devlin2018bert} text enc. (ours) & 1.425 \\
GHOST OpenSeg w. closed vocab. scene enc. \cite{dai2017scannet,wang2022humanise} (ours) & 1.021 \\
GHOST OpenSeg w. $\lambda_{bbox}=0$ (ours) & 1.011 \\
GHOST OpenSeg w. $\lambda_{class}=0$ (ours) & 0.982 \\
GHOST OpenSeg w. $\lambda_{class}=0.1$ (ours) & 0.995 \\
GHOST OpenSeg w. $\lambda_{class}=1.0$ (ours) & 0.970 \\
GHOST OpenSeg (ours) & \textbf{0.952} \\
\bottomrule
\end{tabular}
\label{tab:quant_ablation}
\end{table}

Our ablation study for the walk action subset (\cref{tab:quant_ablation}) highlighted the significance of four components in our method.
The most substantial impact was observed when changing the text and scene encoders, emphasizing the importance of aligning these modalities for improved grounding.
Our two regularization losses showed less impact but remained significant.

\subsection{Qualitative Results}
\cref{fig:fancy}, \cref{fig:qual} and \cref{fig:qual2} present qualitative results for motion generation.
To improve visual fidelity, we display the scene meshes instead of the sampled point clouds.

\begin{table*}[!th]
\centering
\captionof{figure}{Qualitative generation results of the agnostic all-actions models on the HUMANISE dataset.
We display 6 samples for each text, with 3 generated by each model.
Ground truth goal objects are highlighted in red, and accompanying attention maps are depicted with purple camera frustums.
Our GHOST model places the character significantly closer to the goal than the HUMANISE cVAE baseline.}
\resizebox{\textwidth}{!}{
\begin{tabular}{@{}c@{}c@{}c@{}c@{}}
\toprule
\multicolumn{2}{c}{walk to the bed that is far from the recycling bin}
& \multicolumn{2}{c}{sit on the chair that is beside the keyboard} \\
\cmidrule(lr){1-2}
\cmidrule(lr){3-4}
HUMANISE cVAE \cite{wang2022humanise} & GHOST OpenSeg (ours) & HUMANISE cVAE \cite{wang2022humanise} & GHOST OpenSeg (ours) \\
\cmidrule(lr){1-1}
\cmidrule(lr){2-2}
\cmidrule(lr){3-3}
\cmidrule(lr){4-4}
\includegraphics[width=0.239\linewidth]{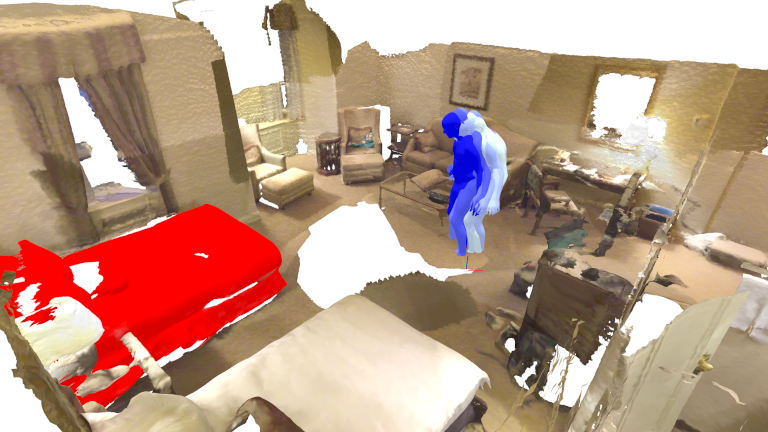} & \includegraphics[width=0.239\linewidth]{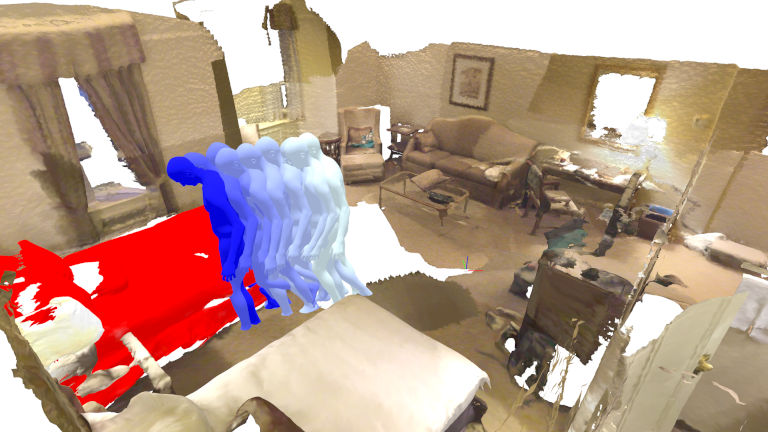} & \includegraphics[width=0.239\linewidth]{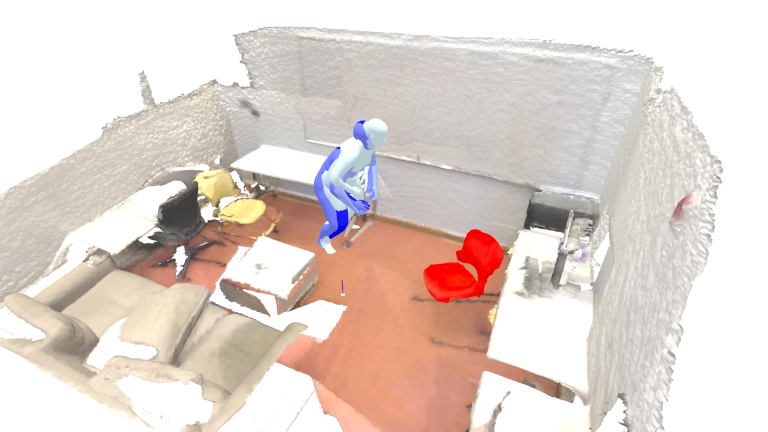} & \includegraphics[width=0.239\linewidth]{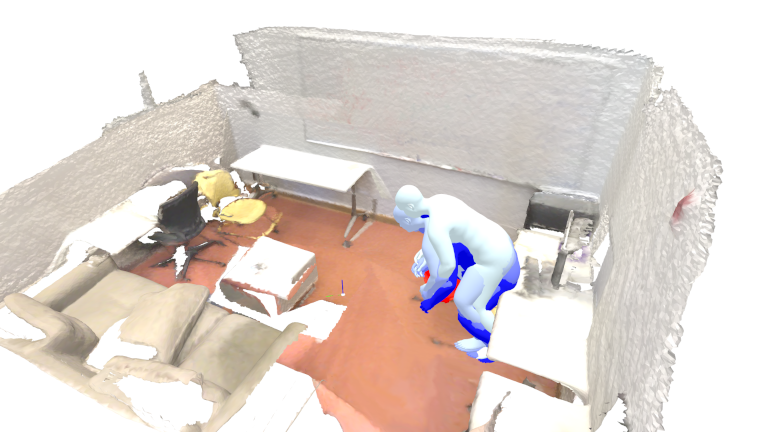}
\\
\includegraphics[width=0.239\linewidth]{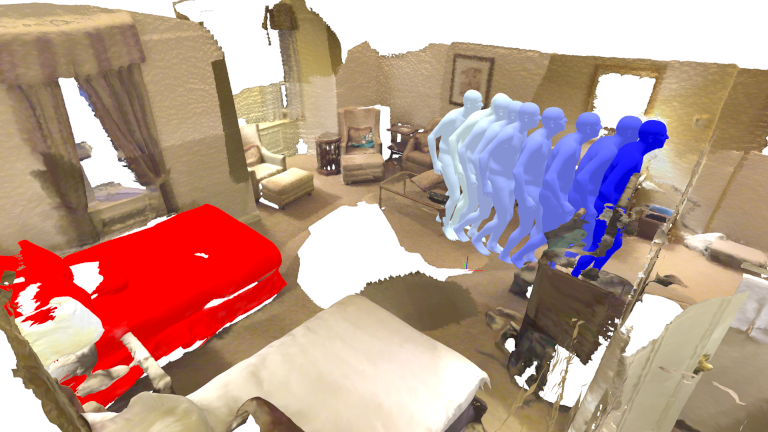} & \includegraphics[width=0.239\linewidth]{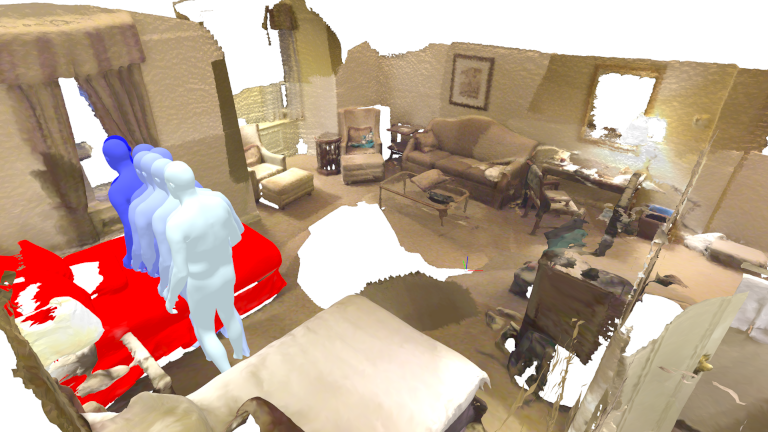} & \includegraphics[width=0.239\linewidth]{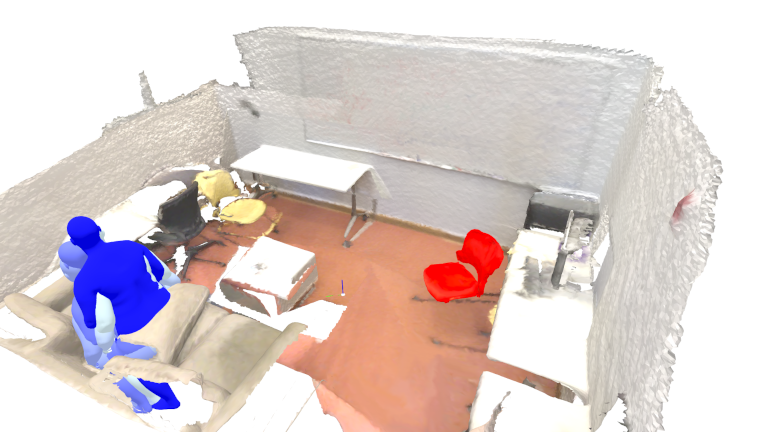} & \includegraphics[width=0.239\linewidth]{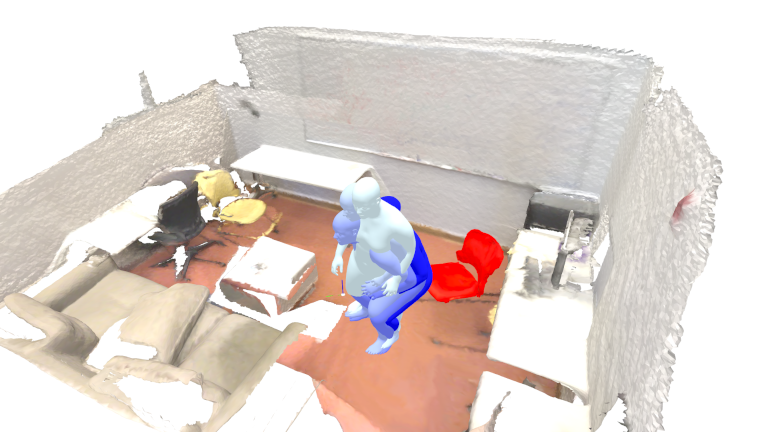}
\\
\includegraphics[width=0.239\linewidth]{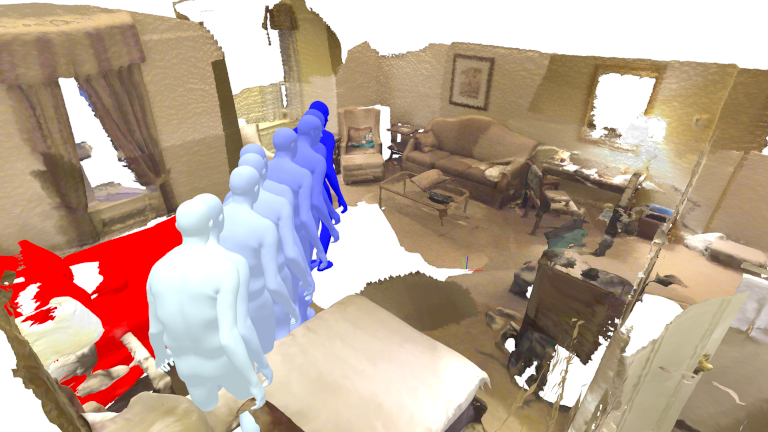} & \includegraphics[width=0.239\linewidth]{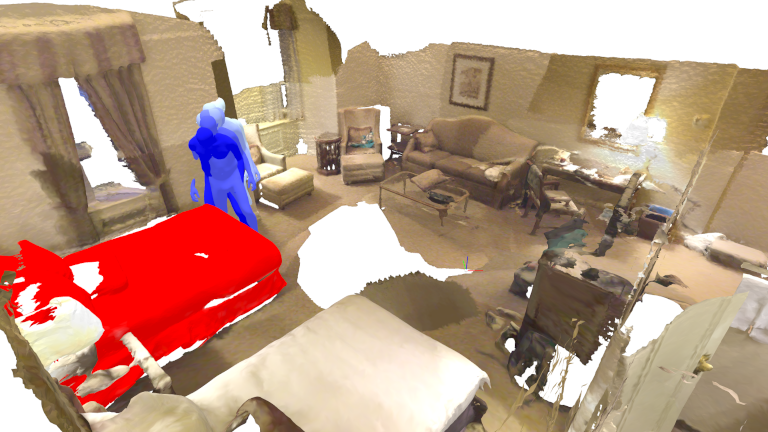} & \includegraphics[width=0.239\linewidth]{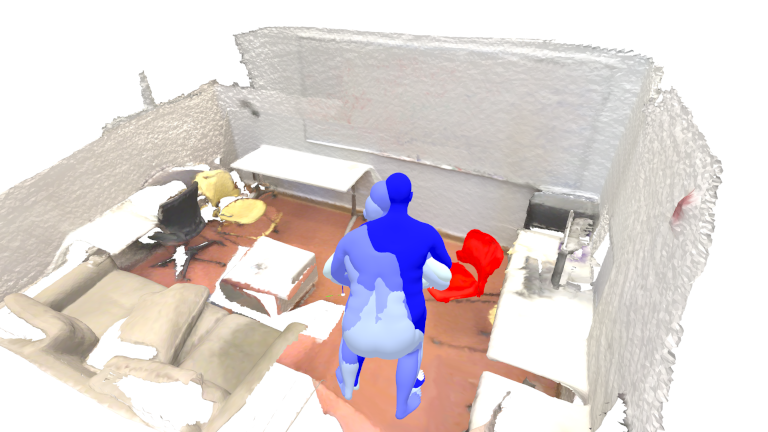} & \includegraphics[width=0.239\linewidth]{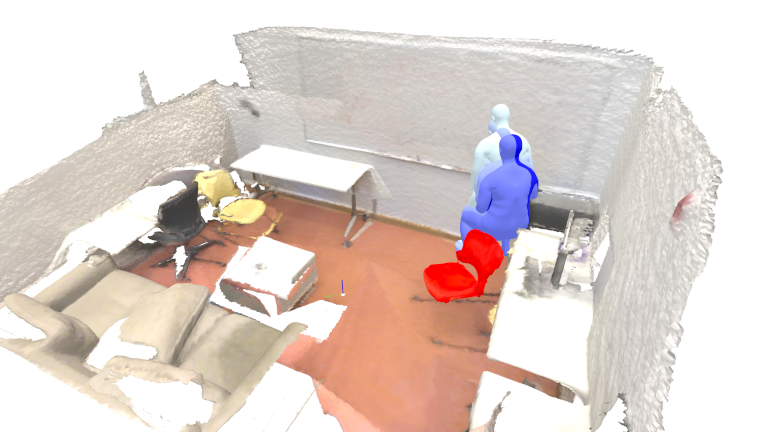} 
\\
\includegraphics[height=0.12825\linewidth,width=0.239\linewidth,keepaspectratio]{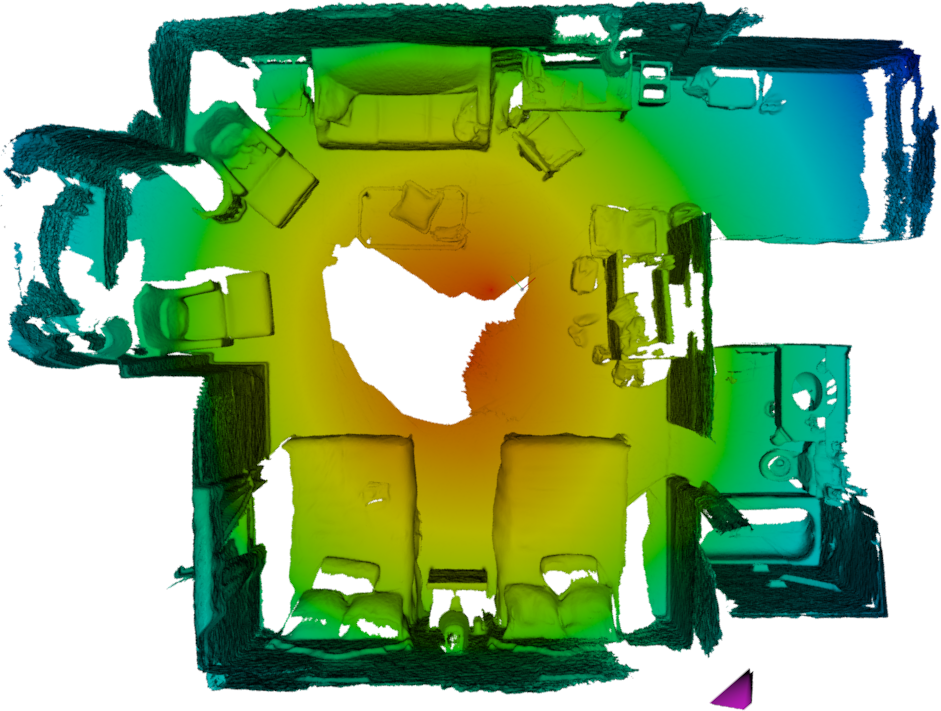} & \includegraphics[height=0.12825\linewidth,width=0.239\linewidth,keepaspectratio]{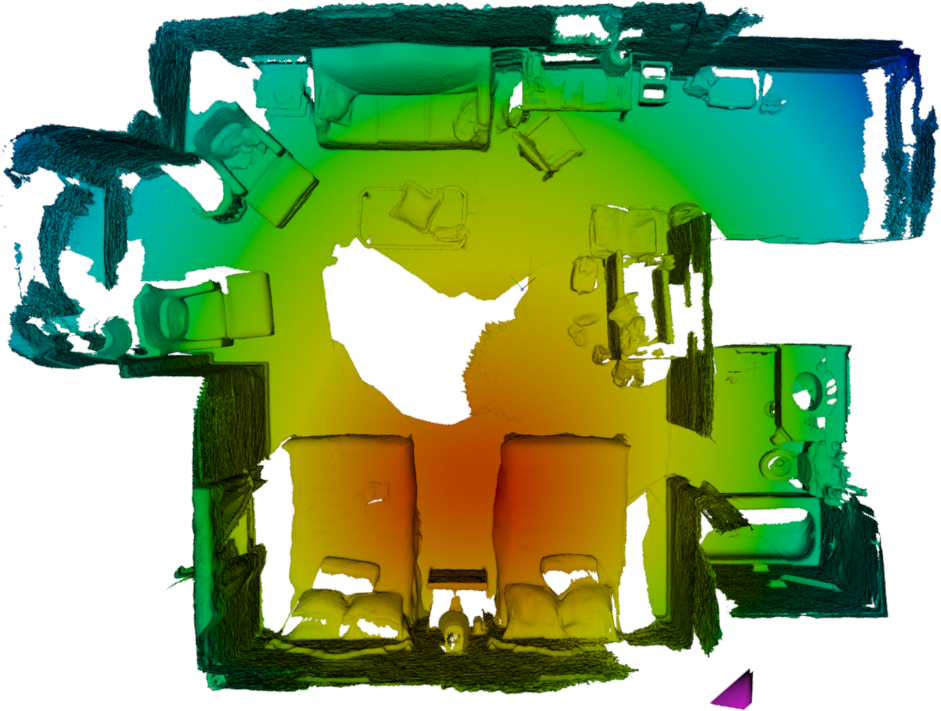} & \includegraphics[height=0.12825\linewidth,width=0.239\linewidth,keepaspectratio]{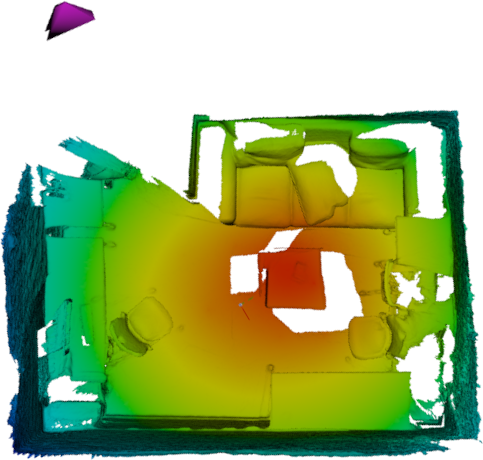} & \includegraphics[height=0.12825\linewidth,width=0.239\linewidth,keepaspectratio]{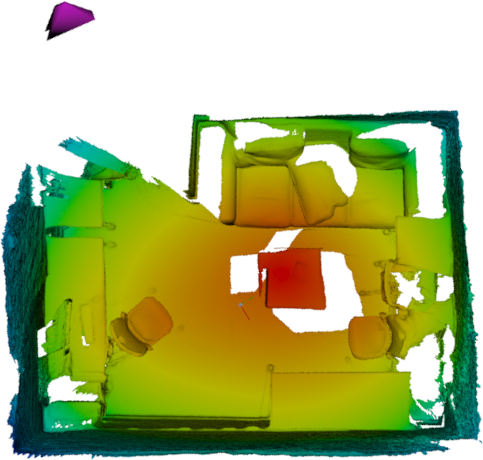}
\\
\midrule
\multicolumn{2}{c}{stand up from the sofa chair} &
\multicolumn{2}{c}{lie on the table that is far away from the cabinet} \\
\cmidrule(lr){1-2}
\cmidrule(lr){3-4}
HUMANISE cVAE \cite{wang2022humanise} & GHOST OpenSeg (ours) & HUMANISE cVAE \cite{wang2022humanise} & GHOST OpenSeg (ours) \\
\cmidrule(lr){1-1}
\cmidrule(lr){2-2}
\cmidrule(lr){3-3}
\cmidrule(lr){4-4}
\includegraphics[width=0.239\linewidth]{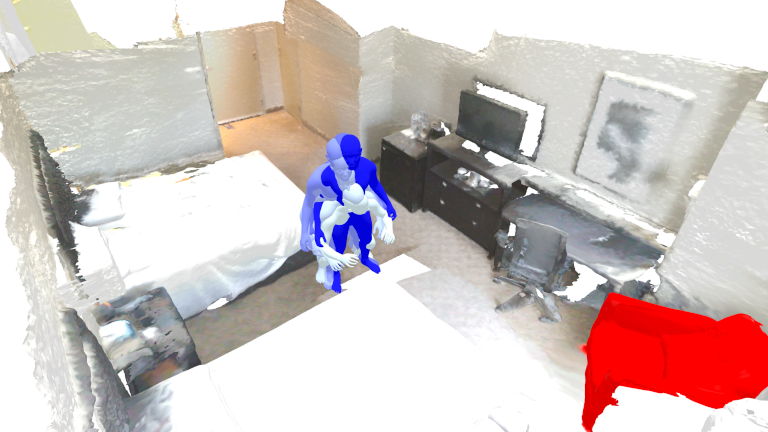} & \includegraphics[width=0.239\linewidth]{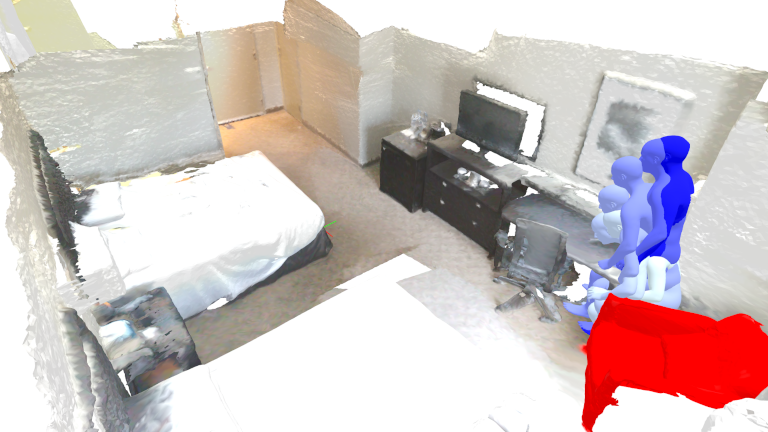} & \includegraphics[width=0.239\linewidth]{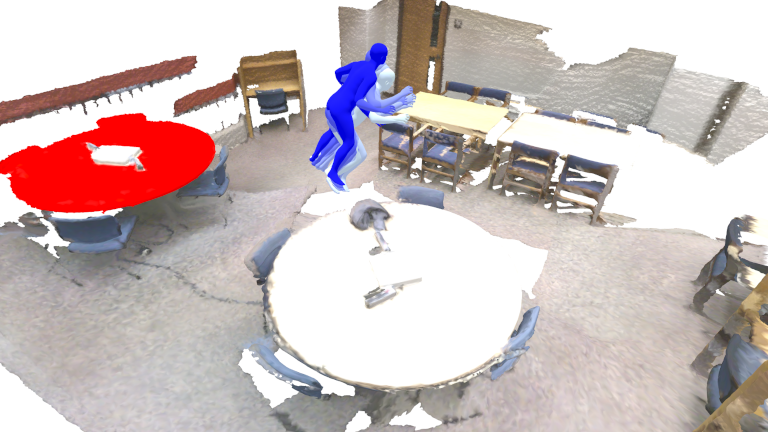} & \includegraphics[width=0.239\linewidth]{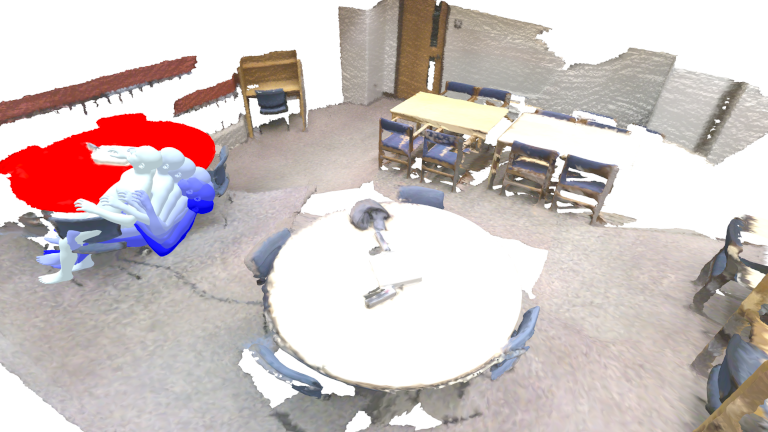}
\\
\includegraphics[width=0.239\linewidth]{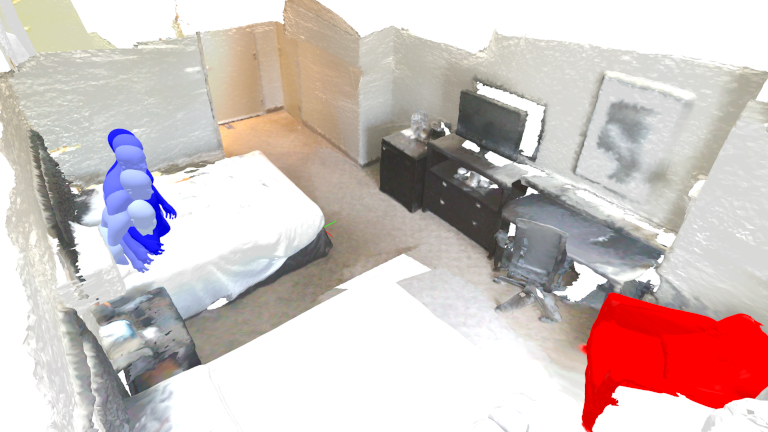} & \includegraphics[width=0.239\linewidth]{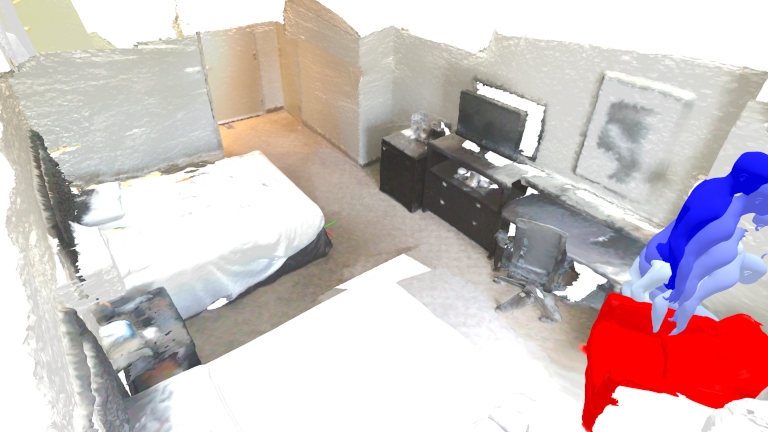} & \includegraphics[width=0.239\linewidth]{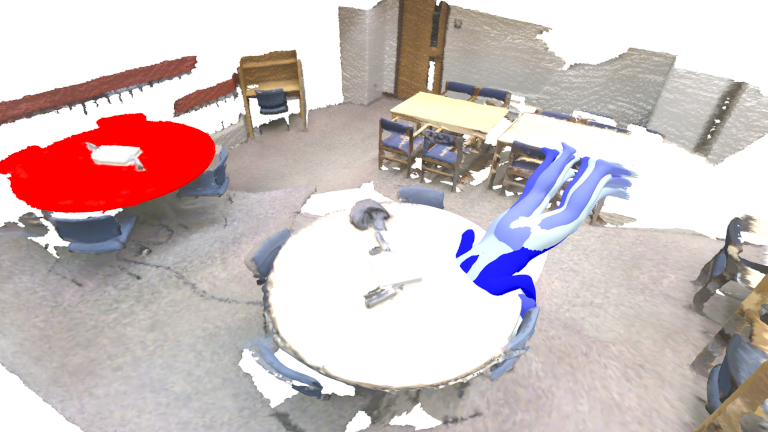} & \includegraphics[width=0.239\linewidth]{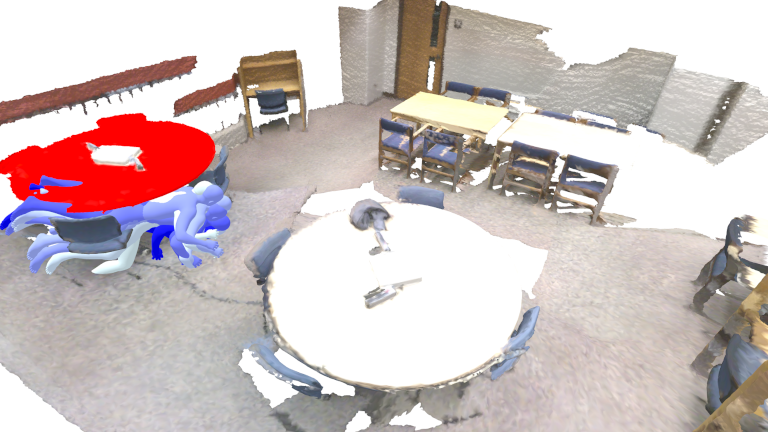}
\\
\includegraphics[height=0.12825\linewidth,width=0.239\linewidth,keepaspectratio]{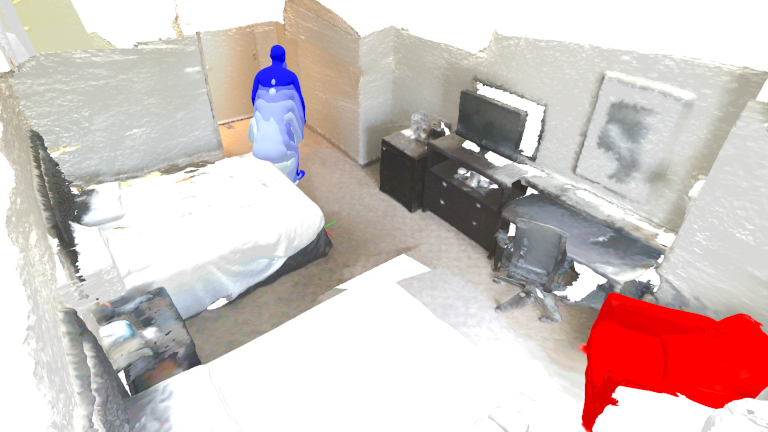} & \includegraphics[height=0.12825\linewidth,width=0.239\linewidth,keepaspectratio]{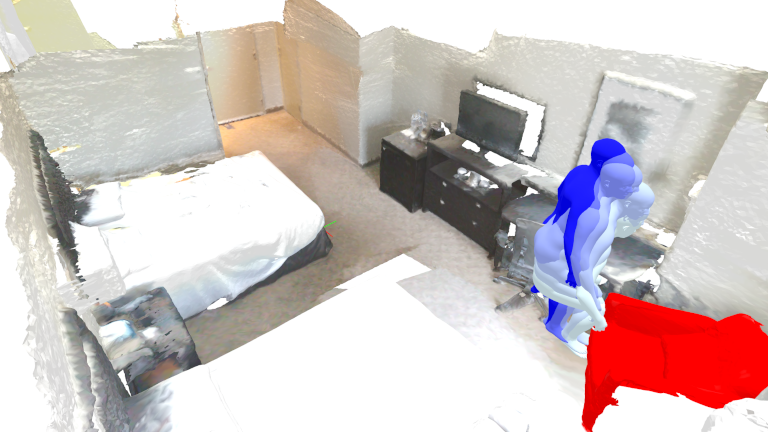} & \includegraphics[height=0.12825\linewidth,width=0.239\linewidth,keepaspectratio]{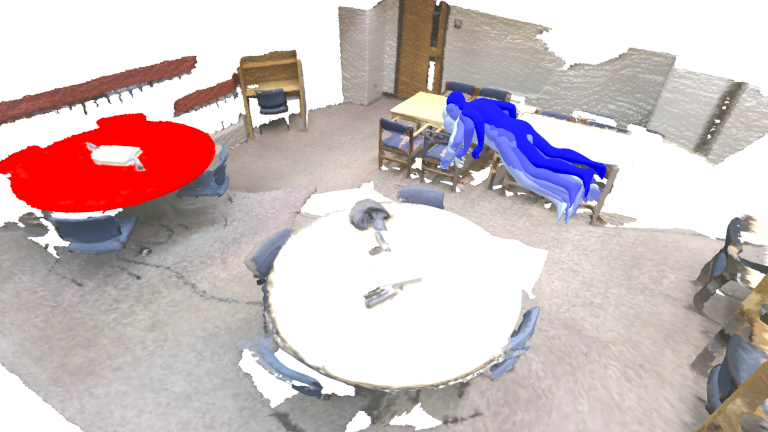} & \includegraphics[height=0.12825\linewidth,width=0.239\linewidth,keepaspectratio]{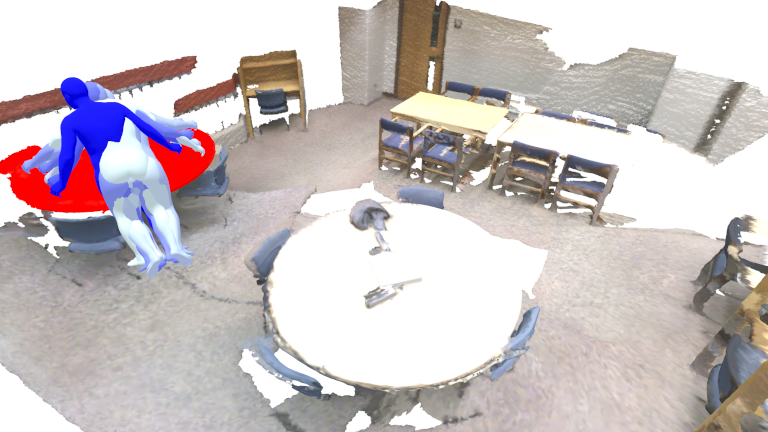}
\\
\includegraphics[height=0.12825\linewidth,width=0.239\linewidth,keepaspectratio]{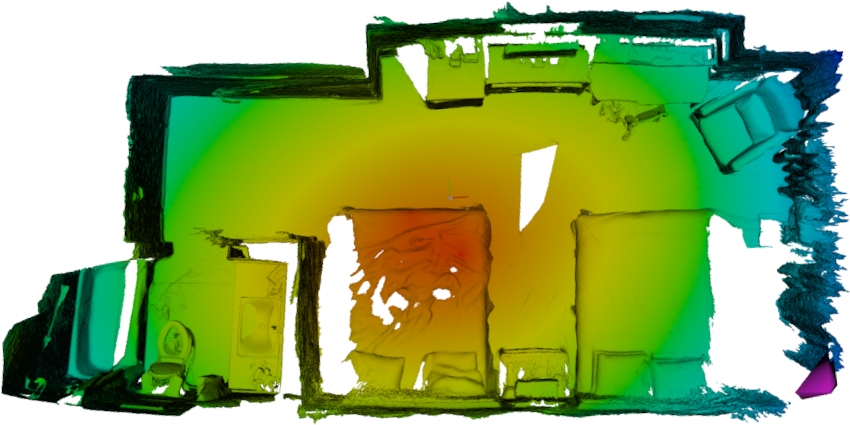} & \includegraphics[height=0.12825\linewidth,width=0.239\linewidth,keepaspectratio]{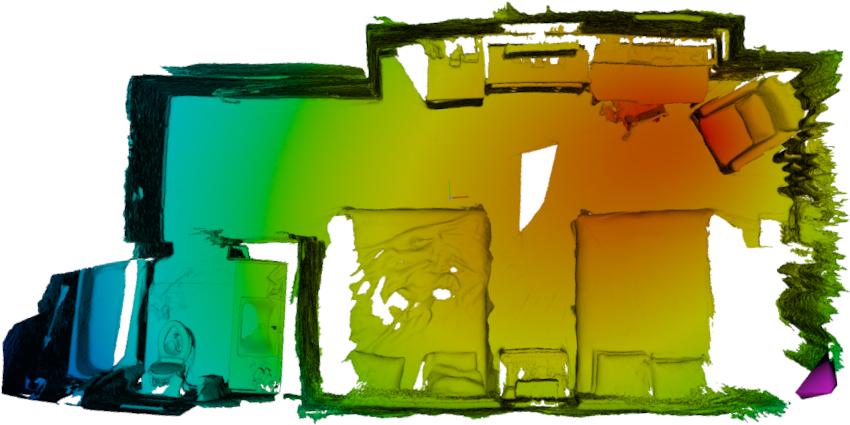} & \includegraphics[height=0.12825\linewidth,width=0.239\linewidth,keepaspectratio]{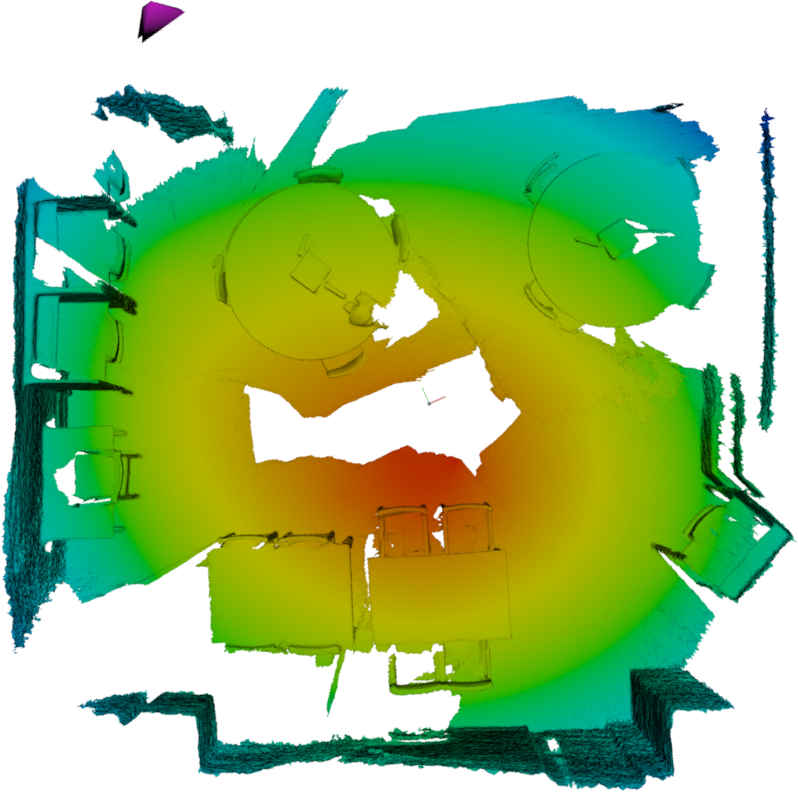} & \includegraphics[height=0.12825\linewidth,width=0.239\linewidth,keepaspectratio]{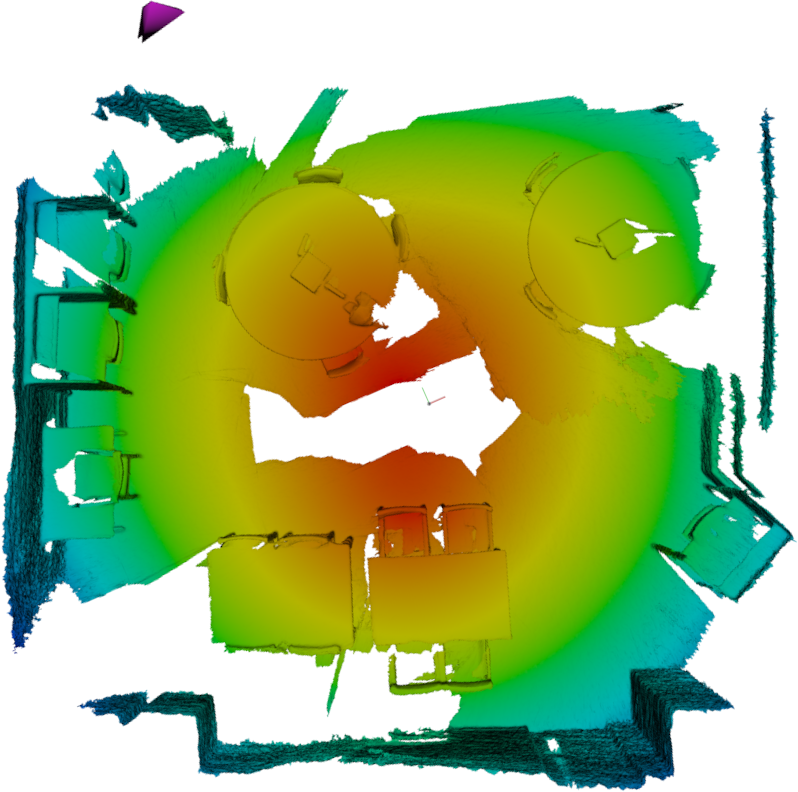}
\\
\bottomrule
\end{tabular}
}
\label{fig:qual}
\end{table*}

In \cref{fig:fancy} and \cref{fig:qual}, we compare our method against the HUMANISE cVAE baseline for all four motions.
\cref{fig:fancy} focuses on recognizing multiple objects within a single scene, while \cref{fig:qual} illustrates generalization across scenes and the randomness of multiple samples per scene.
Therefore, in \cref{fig:qual}, we also highlight the ground truth goal object in red color, and show the corresponding attention maps.
It can be observed that the HUMANISE cVAE fails to accurately locate the goal object in the scene, and the human location tends to be biased towards the center of the scene (as confirmed by the \cref{fig:qual} attention maps).
On the other hand, our GHOST OpenSeg model successfully identifies the goal object, places the human close enough, and generates plausible interactions.
However, it is worth noting that not every sampled motion has a correct fine-grained orientation, the attention maps may still be biased towards the center, and some motions may exhibit scene penetrations.

We present qualitative results for ablation in \cref{fig:qual2}, comparing our full model against its variants without the proposed regularization losses.
These qualitative findings align with the numbers presented in \cref{tab:quant_ablation}, but show more significance for both regularizers.
Notably, the bounding box loss seems particularly crucial, emphasizing the importance of localizing the target object by inferring its size.

\begin{table*}[!th]
\centering
\captionof{figure}{Qualitative generation results of ablation on the walk action subset of the HUMANISE dataset.
We display 3 samples for the same text, with 1 generated by each model.
Ground truth goal object is highlighted in red.
Our GHOST model places the character significantly closer to the goal with our proposed regularization losses.}
\begin{tabular}{@{}ccc@{}}
\toprule
\multicolumn{3}{c}{walk to the chair that is farthest from the end table} \\
\cmidrule(lr){1-3}
GHOST OpenSeg & GHOST OpenSeg & GHOST OpenSeg \\
w. $\lambda_{bbox}=0$ (ours) & w. $\lambda_{class}=0$ (ours) & (ours) \\
\cmidrule(lr){1-1}
\cmidrule(lr){2-2}
\cmidrule(lr){3-3}
\includegraphics[width=0.239\linewidth]{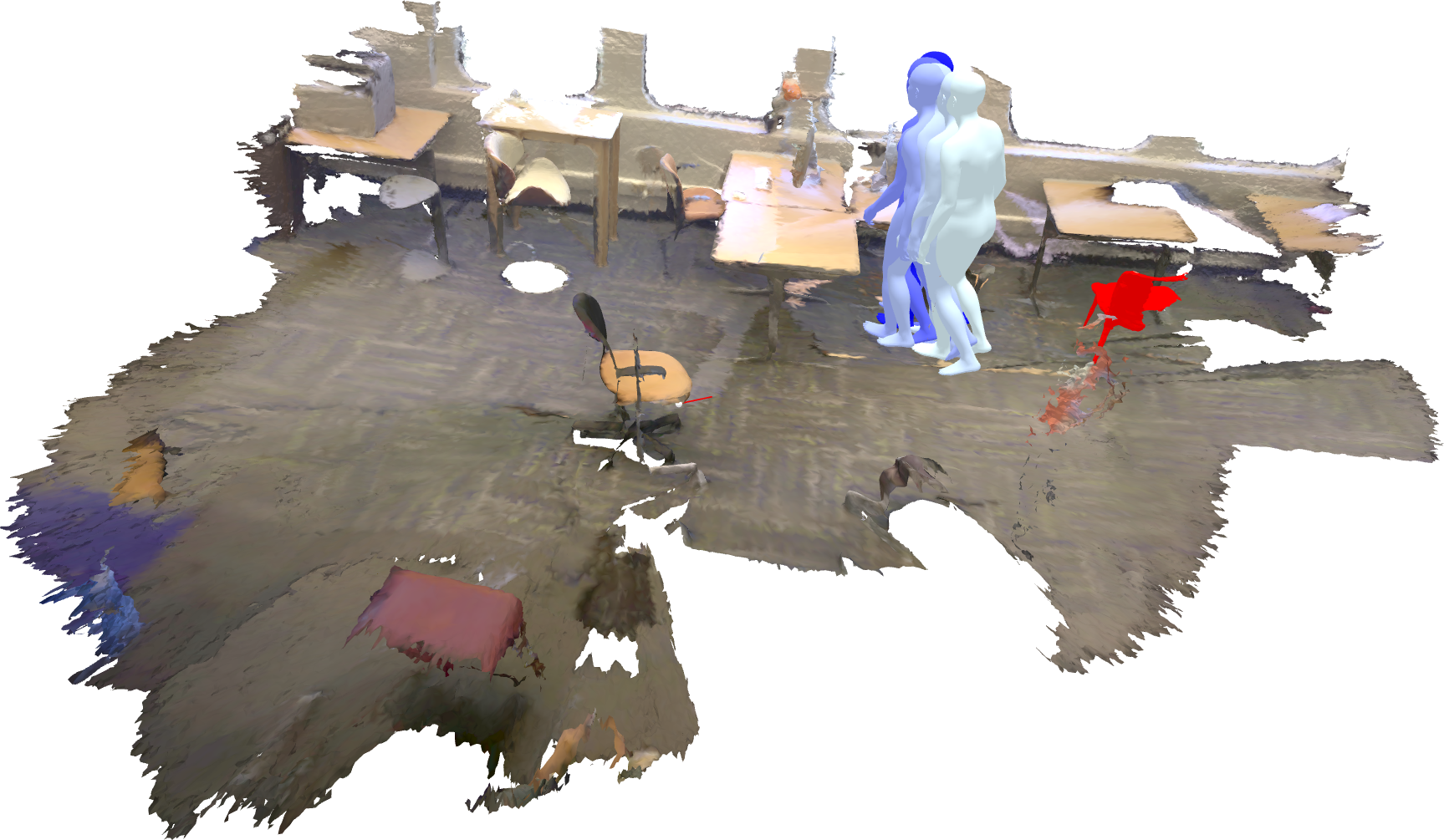} & \includegraphics[width=0.239\linewidth]{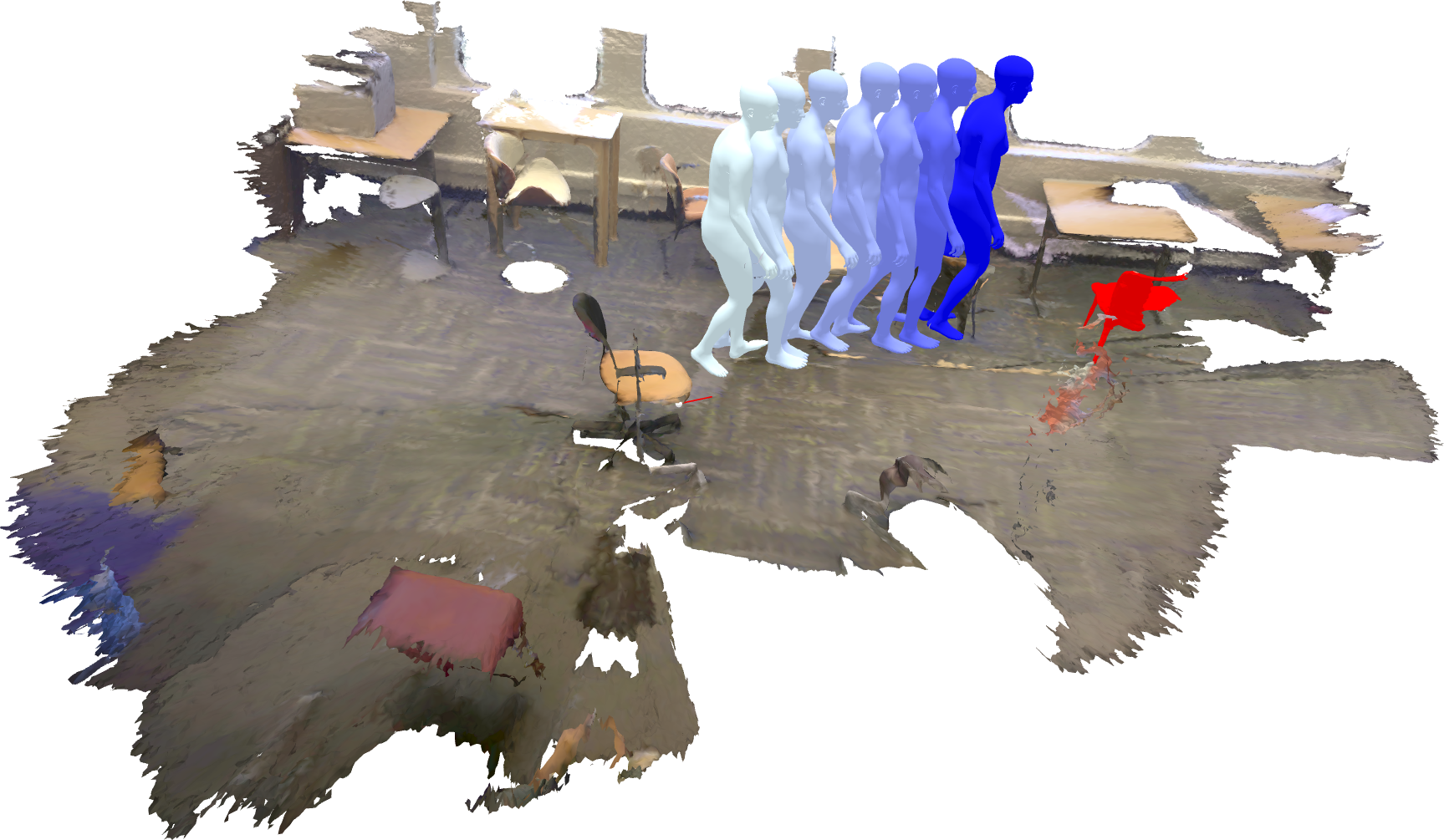} & \includegraphics[width=0.239\linewidth]{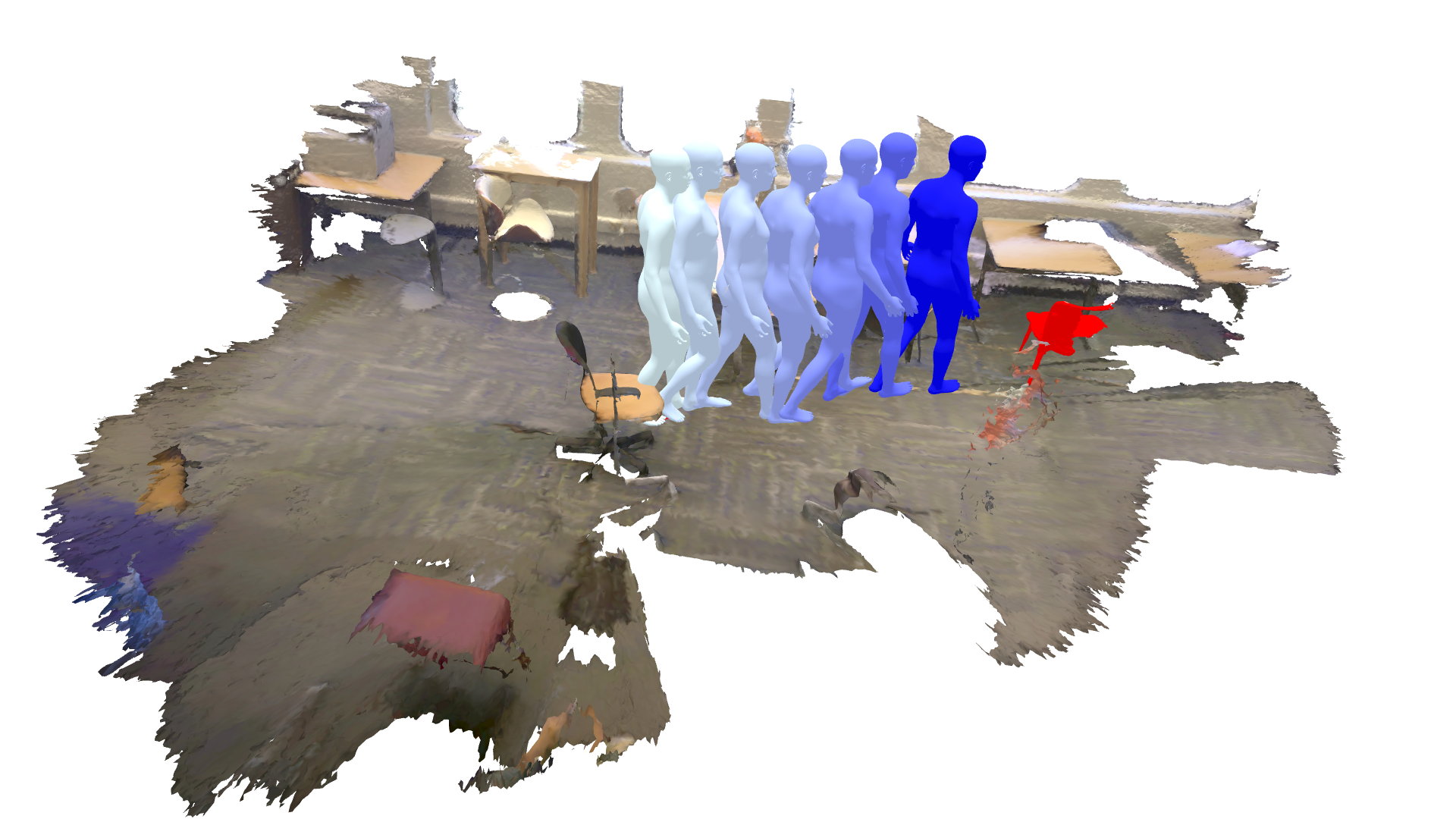}
\\
\bottomrule
\end{tabular}
\label{fig:qual2}
\end{table*}

\subsection{Computational Analysis}
We report parameter counts as dot sizes in \cref{fig:fancyc} (depending on variant, $1.5\times$ to $3.9\times$ larger than HUMANISE cVAE).
Specifically, our scene encoder is $1.6\times$ larger, and outputs a $16\times$ to $24\times$ larger representation.
Yet, motion sampling takes only $1.3\times$ longer (wall-clock time $\approx\SI{0.19}{\second}$ on A100 GPU).

\section{Conclusion}

In this paper, we introduced GHOST, a novel text-and-scene-conditional human motion generation framework.
Our approach is designed to enhance text-scene grounding and motion placement by utilizing open vocabulary knowledge distillation for CLIP space alignment and incorporating additional regularization.
Quantitatively, all three implementations of our framework significantly outperformed the HUMANISE cVAE baseline, and our best model exhibited superior qualitative performance as well.

Nonetheless, our current solution has some limitations.
It still exhibits goal identification, orientation and scene penetration errors, suggesting the necessity for better VLMs, teacher models and further regularization.
Future directions may involve substituting the cVAE with a diffusion model, extending grounding to non-goal/non-anchor objects, post-processing our results with contact optimization, as well as addressing generalization to natural texts and more actions.

In summary, our work advances the localization aspect of human motion generation.
We anticipate that our findings will catalyze further research in this direction, driving the development of even more sophisticated and accurate techniques with profound practical implications.

%
%
\bibliographystyle{splncs04}
\bibliography{main}
\end{document}


\title{GHOST: \underline{G}rounded \underline{H}uman Motion Generation with \underline{O}pen Vocabulary \underline{S}cene-and-\underline{T}ext Contexts} 

\titlerunning{GHOST: \underline{G}rounded \underline{H}uman Motion Generation}

\author{Zolt\'{a}n~\'{A}.~Milacski\inst{1}\orcidlink{0000-0002-3135-2936} \and
Koichiro~Niinuma\inst{2}\orcidlink{0000-0001-8367-3988} \and
Ryosuke~Kawamura\inst{2}\orcidlink{0000-0001-5133-9838} \and
Fernando~de~la~Torre\inst{1}\orcidlink{0009-0007-3760-3007} \and
L\'{a}szl\'{o}~A.~Jeni\inst{1}\orcidlink{0000-0002-2830-700X}}

\authorrunning{Z.~\'{A}.~Milacski, K.~Niinuma et al.}

\institute{Robotics Institute, Carnegie Mellon University, Pittsburgh PA, USA 
\email{\{zmilacsk@andrew,ftorre@cs,laszlojeni@\}.cmu.edu}
\and
Fujitsu Research of America, Pittsburgh PA, USA
\email{\{kniinuma,k.ryosuke\}@fujitsu.com}}

\title{GHOST: \underline{G}rounded \underline{H}uman Motion Generation with \underline{O}pen Vocabulary \underline{S}cene-and-\underline{T}ext Contexts\\
\emph{Supplementary Material}}  

\maketitle
\thispagestyle{empty}
\appendix

\section{Additional Experiments}



\subsection{Additional Evaluation Metrics}
In this section, we provide additional performance metrics.
While the main paper primarily emphasizes the distance between the generated motions and the goal object, these metrics offer additional insights into those results.

\paragraph{Condition.}
Here, we compute 2 metrics for evaluating the grounding performance of the condition module.
First, we calculate the cosine similarity of the encodings of the text prompt and the pooled cloud point that is nearest to the goal object center:
\begin{equation}
    \cos\left(\mathcal{E}^{text}(\bm{L}),Pool\left(\mathcal{E}^{3D}(\bm{S})\right)_{goal\ center,4:}\right),
\end{equation}
where $\mathcal{E}^{text}$ is the text encoder, and $Pool$ is either identity for the HUMANISE cVAE or the $k$-nearest neighbor downsampling module for our GHOST.
We specifically employ goal object center indexing to ensure a fair comparison between all methods, as some of them lack a ground truth goal object mask at this level.
Second, we report the $\mathcal{L}_{center}$ MSE regularization loss in meters from the main paper for regressing the goal object center point from both input modalities.
We average both metrics across samples.

\paragraph{Reconstruction.}
This task is significantly easier than generation, given the availability of the ground truth motion location as an input.
Yet, we present the respective performance results here in the supplementary material.

We assess the motion reconstruction capability by computing the MAE ($\ell_1$ error) $\times 100$ between the ground truth and predicted SMPL-X parameters, specifically for global translation $\bm{t}$, global orientation $\bm{r}$, and body pose $\bm{\theta}$.
Following \cite{wang2022humanise,wang2021synthesizing} to obtain more interpretable scores, we also calculate the Mean Per Vertex Position Error (MPVPE) and Mean Per Joint Position Error (MPJPE) \cite{ionescu2013human3} in millimeters.
To handle sequences of various lengths, we average these results over the temporal dimension, and finally, across examples.

\paragraph{Generation.}
We report the Average Pairwise Distance (APD) \cite{yuan2020dlow,zhang2021we}, which is the average $\ell_2$ distance between corresponding body markers:
\begin{equation}
\mathrm{APD}(\bm{L}, \bm{S})=
\frac{1}{Q}\sum_{\substack{i=1 \\ j=1 \\ i \neq j}}^{K'}{\sum_{t=1}^{T}{\biggl\|\bm{M} \circ \Bigl(\hat{\bm{\mathcal{M}}}_t^{(i)}}} 
{{{-\hat{\bm{\mathcal{M}}}_t^{(j)}\Bigr)\biggr\|_2}}},
\label{eq:apd}
\end{equation}
where $Q=K'(K'-1)T$; and $\bm{M}\in\{0,1\}^{\SI{10475}{}}$ is a binary mask that selects 67 vertices from the SMPL-X model, corresponding to commonly used motion capture landmarks \cite{zhang2021we}.
We sample $K'=20$ motions here.
While \cite{wang2022humanise} employs APD as a measure of diversity, favoring higher values, it is essential to balance this with the goal object distance $d(\bm{L},\bm{S})$ to avoid favoring overly random predictions.
Therefore, we argue that, given the need for motions distributed around specific goal objects and the substantial distances involved, lower APD values are preferable.

Furthermore, we also present standard deviations corresponding to the average goal object distances $d(\bm{L},\bm{S})$ in the main paper.

\paragraph{Perceptual Study.}
We present comprehensive per-subject results for our perceptual experiment.

\subsection{Additional Quantitative Results}
\cref{tab:quant} collects our more detailed results.
In condition module evaluation, we found that our GHOST methods achieved significantly larger cosine similarities than the HUMANISE cVAE baseline, indicating better text-scene grounding.
However, the goal object center regularization loss $\mathcal{L}_{center}$ correlated the best with the final goal object distance metric.
Interestingly, our GHOST LSeg sometimes outperformed the OpenSeg variant in this regard, but the latter still achieved more reliable results.
In reconstruction, our global orientation and pose errors were competitive, but the global translations, MPVPEs and MPJPEs were superior for the HUMANISE cVAE.
This may be attributed to the impact of our additional regularization losses, which counteract reconstruction efforts, suggesting an area for future enhancement.
In generation, the goal object distance standard deviations were generally comparable, while our APDs were smaller indicating generated motions that are converging more towards the same position.

\begin{table*}[!h]
\centering
\caption{
Quantitative results of reconstruction and generation experiments on the HUMANISE dataset.
The winning numbers are highlighted in bold for each action subset.}
\resizebox{\textwidth}{!}{%
\begin{tabular}{@{}lcccccccccc@{}}
\toprule
& & \multicolumn{2}{c}{Condition} & \multicolumn{5}{c}{Reconstruction} & \multicolumn{2}{c}{Generation} \\
\cmidrule(lr){3-4} \cmidrule(lr){5-9} \cmidrule(lr){10-11}
& & Text-Goal obj. & MSE Goal obj. & \multicolumn{3}{c}{MAE $\times$ 100} & MPVPE & MPJPE & Goal obj. & \\
\cmidrule(lr){5-7}
Action & Method & enc. $\cos$ sim. $\uparrow$ & center reg. ($\SI{}{m}$) $\downarrow$ & trans. $\downarrow$ & orient. $\downarrow$ & pose $\downarrow$ &($\SI{}{\milli\metre}$) $\downarrow$ & ($\SI{}{\milli\metre}$) $\downarrow$ & dist.$\pm$std ($\SI{}{m}$) $\downarrow$ & APD $\downarrow$ \\
\midrule
\multirow{4}{*}[0em]{walk} & HUMANISE cVAE \cite{wang2022humanise} & $\phantom{0}$1.75 & 1.372 & 5.84 & 2.80 & 1.85 & 123.88 & 125.05 & 1.370$\pm$0.839 & 12.83 \\
& GHOST LSeg (ours) & $\phantom{0}$9.16 & 1.090 & 6.17	& 2.64 & 1.83 & 128.59 & 129.59 & 1.090$\pm$0.891 & 10.96 \\
& GHOST OpenSeg (ours) & $\phantom{0}$5.08 & \textbf{0.990} & 5.97 & 2.86 & 1.90 & 126.66 & 128.02 & \textbf{0.952}$\pm$0.919 & 10.97 \\
& GHOST OVSeg (ours) & \textbf{10.25} & 1.101 & 6.45 & 2.88 & 1.87 & 137.38 & 138.43 & 1.027$\pm$0.945 & \textbf{10.62} \\
\midrule
\multirow{4}{*}[0em]{sit} & HUMANISE cVAE \cite{wang2022humanise} & $\phantom{0}$1.06 & 0.910 & 5.17 & 3.19 & 1.77 & 112.43 & 113.28 & 0.903$\pm$0.744 & 10.12 \\
& GHOST LSeg (ours) & 10.16 & \textbf{0.621} & 6.00 & 2.89 & 1.74 & 127.64 & 128.48 & 0.695$\pm$0.655 & $\phantom{0}$9.28 \\
& GHOST OpenSeg (ours) & $\phantom{0}$6.97 & 0.709 & 5.92 & 2.96 & 1.79 & 125.41 & 126.10 & \textbf{0.668}$\pm$0.708 & $\phantom{0}$8.59 \\
& GHOST OVSeg (ours) & \textbf{12.40} & 0.735 & 6.10 & 3.17 & 1.77 & 129.72 & 130.37 & 0.680$\pm$0.743 & $\phantom{0}$\textbf{8.29} \\
\midrule
\multirow{4}{*}[0em]{stand up} & HUMANISE cVAE \cite{wang2022humanise} & -0.18 & 0.875 & 5.63 & 3.43 & 1.69 & 124.84 & 126.05 & 0.802$\pm$0.711 & $\phantom{0}$9.57 \\
& GHOST LSeg (ours) & 12.04 & 0.861 & 6.09 & 3.51 & 1.71 & 130.60 & 131.71 & 0.767$\pm$0.742 & $\phantom{0}$8.89 \\
& GHOST OpenSeg (ours) & $\phantom{0}$6.52 & \textbf{0.595} & 6.32 & 3.73 & 1.76 & 134.62 & 135.70 & \textbf{0.600}$\pm$0.600 & $\phantom{0}$\textbf{8.45} \\
& GHOST OVSeg (ours) & \textbf{13.22} & 0.674 & 6.91 & 3.58 & 1.74 & 148.29 & 149.25 & 0.626$\pm$0.681 & $\phantom{0}$8.59 \\
\midrule
\multirow{4}{*}[0em]{lie} & HUMANISE cVAE \cite{wang2022humanise} & -3.64 & 0.397 & 6.46 & 3.09 & 0.76 & 136.20 & 136.87 & 0.196$\pm$0.476 & $\phantom{0}$9.18 \\
& GHOST LSeg (ours) & \textbf{13.91} & \textbf{0.327} & 7.84 & 3.04 & 0.76 & 169.87 & 170.54 & \textbf{0.185}$\pm$0.425 & $\phantom{0}$8.87 \\
& GHOST OpenSeg (ours) & $\phantom{0}$4.83 & 0.410 & 6.99 & 3.01 & 0.88 & 150.64 & 151.45 & 0.200$\pm$0.468 & $\phantom{0}$\textbf{8.54} \\
& GHOST OVSeg (ours) & 10.59 & 0.623 & 6.95 & 3.22 & 0.83 & 148.60 & 149.70 & 0.263$\pm$0.603 & $\phantom{0}$8.97 \\
\midrule
\multirow{4}{*}[0em]{all} & HUMANISE cVAE \cite{wang2022humanise} & $\phantom{0}$4.84 & 1.044 & 4.20 & 2.91 & 1.96 & $\phantom{0}$96.53 & $\phantom{0}$98.01 & 1.008$\pm$0.838 & 11.83 \\
& GHOST LSeg (ours) & $\phantom{0}$9.49 & \textbf{0.754} & 4.37 & 2.87 & 1.91 & $\phantom{0}$98.62 & $\phantom{0}$99.93 & 0.748$\pm$0.810 & $\phantom{0}$\textbf{9.54} \\
& GHOST OpenSeg (ours) & $\phantom{0}$6.17 & 0.788 & 4.37 & 2.82 & 1.93 & $\phantom{0}$98.76 & 100.02 & \textbf{0.732}$\pm$0.837 & $\phantom{0}$9.80 \\
& GHOST OVSeg (ours) & \textbf{10.54} & 0.823 & 4.08 & 2.92 & 1.90 & $\phantom{0}$93.15 & $\phantom{0}$94.54 & 0.767$\pm$0.829 & 10.08 \\
\bottomrule
\end{tabular}%
}
\label{tab:quant}
\end{table*}

\begin{table*}[!h]
\centering
\caption{
Quantitative results of ablation experiments on the walk action in the HUMANISE dataset.
The winning numbers are highlighted in bold.}
\resizebox{\textwidth}{!}{%
\begin{tabular}{@{}lccccccccc@{}}
\toprule
& \multicolumn{2}{c}{Condition} & \multicolumn{5}{c}{Reconstruction} & \multicolumn{2}{c}{Generation} \\
\cmidrule(lr){2-3} \cmidrule{4-8} \cmidrule(lr){9-10}
& Text-Goal obj. & MSE Goal obj. & \multicolumn{3}{c}{MAE $\times$ 100} & MPVPE & MPJPE & Goal obj. \\
\cmidrule(lr){4-6}
Method & enc. $\cos$ sim. $\uparrow$ & center reg. ($\SI{}{m}$) $\downarrow$ & trans. $\downarrow$ & orient. $\downarrow$ & pose $\downarrow$ &($\SI{}{\milli\metre}$) $\downarrow$ & ($\SI{}{\milli\metre}$) $\downarrow$ & dist.$\pm$std ($\SI{}{m}$) $\downarrow$ & APD $\downarrow$ \\
\midrule
GHOST OpenSeg w. BERT \cite{devlin2018bert} text enc. (ours) & 3.04 & 1.574 & 5.85 & 3.02 & 1.93 & 124.71 & 125.98 & 1.425$\pm$0.917 & 11.28 \\
GHOST OpenSeg w. closed vocab. scene enc. \cite{dai2017scannet,wang2022humanise} (ours) & \textbf{7.81} & 1.230 & 5.95 & 2.96 & 1.88 & 125.35 & 126.53 & 1.021$\pm$1.032 & \textbf{10.38} \\
GHOST OpenSeg w. $\lambda_{bbox}=0$ (ours) & 4.96 & \textbf{0.990} & 5.92 & 2.82 & 1.90 & 125.43 & 126.70 & 1.011$\pm$0.860 & 11.65 \\
GHOST OpenSeg w. $\lambda_{class}=0$ (ours) & 4.91 & 1.028 & 6.07 & 2.75 & 1.87 & 128.67 & 129.85 & 0.982$\pm$0.925 & 11.09 \\
GHOST OpenSeg (ours) & 5.08 & \textbf{0.990} & 5.97 & 2.86 & 1.90 & 126.66 & 128.02 & \textbf{0.952}$\pm$0.919 & 10.97\\
\bottomrule
\end{tabular}%
}
\label{tab:quant_ablation}
\end{table*}

\begin{table*}[!h]
\centering
\caption{
Quantitative results of the perceptual study of agnostic all-actions models trained on the entire HUMANISE dataset.
The winning numbers are highlighted in bold.}

\resizebox{\textwidth}{!}{%
\begin{tabular}{@{}lccccccccc@{}}
\toprule
& \multicolumn{9}{c}{Frequency of User Preference $\uparrow$}\\
\cmidrule(lr){2-10}
Method & User 1 & User 2 & User 3 & User 4 & User 5 & User 6 & User 7 & User 8 & User 9\\
\midrule
HUMANISE cVAE \cite{wang2022humanise} & 15 & 22 & 17 & 22 & 22 & 21 & 27 & 21 & 26\\ 
GHOST OpenSeg (ours) & \textbf{45} & \textbf{38} & \textbf{43} & \textbf{38} & \textbf{38} & \textbf{39} & \textbf{33} & \textbf{39} & \textbf{34} \\
\midrule
& User 10 & User 11 & User 12 & User 13 & User 14 & User 15 & User 16 & User 17 & User 18\\
\midrule
HUMANISE cVAE \cite{wang2022humanise} & 23 & 27 & 20 & 27 & 21 & 22 & 21 & 25 & 28\\
GHOST OpenSeg (ours) & \textbf{37} & \textbf{33} & \textbf{40} & \textbf{33} & \textbf{39} & \textbf{38} & \textbf{39} & \textbf{35} & \textbf{32} \\
\midrule
& User 19 & User 20 & User 21 & User 22 & User 23 & User 24 & User 25 & User 26 & User 27\\
\midrule
HUMANISE cVAE \cite{wang2022humanise} & 23 & 21 & 21 & 20 & 20 & 20 & 24 & 20 & 19\\
GHOST OpenSeg (ours) & \textbf{37} & \textbf{39} & \textbf{39} & \textbf{40} & \textbf{40} & \textbf{40} & \textbf{36} & \textbf{40} & \textbf{41} \\
\bottomrule
\end{tabular}%
}
\label{tab:quant_user}
\end{table*}

\cref{tab:quant_ablation} shows the corresponding numbers for ablation.
We observe that employing a closed vocabulary scene encoder resulted in strong text-goal cosine similarity and APD scores.
However, it still struggled to accurately regress the center of the goal object, potentially due to ambiguities between the embeddings of the goal object and the rest of the 3D scene.
As expected, our regularization losses sometimes hampered reconstruction.

\cref{tab:quant_user} details our perceptual study results.
All $27$ subjects picked the samples generated by our GHOST method more frequently, with preferences up to $75\%$.

%
%
\bibliographystyle{splncs04}
\bibliography{main}